\documentclass[review]{elsarticle}
\usepackage{graphicx}
\usepackage{amsmath}
\usepackage{multirow}
\usepackage{color}
\usepackage{epstopdf}
\usepackage{lineno,hyperref}
\modulolinenumbers[5]
\usepackage{xspace}
\usepackage{amsmath}
\usepackage{amsfonts}
\usepackage{geometry}
\usepackage{soul}
\journal{}

\biboptions{authoryear}

\begin{document}

\begin{frontmatter}

\title{CSRNet: Cascaded Selective Resolution Network for Real-time Semantic Segmentation}


\author[mymainaddress]{Jing-Jing Xiong\corref{mycorrespondingauthor}}
\cortext[mycorrespondingauthor]{Corresponding author}
\ead[url]{jingxiong9-c@my.cityu.edu.hk}
\author[mymainaddress]{Lai-Man Po}
\ead[url]{eelmpo@cityu.edu.hk}

\author[mymainaddress]{Wing-Yin Yu}
\ead[url]{wingyinyu8-c@my.cityu.edu.hk}

\author[mymainaddress]{Chang Zhou}
\ead[url]{chanzhou3-c@my.cityu.edu.hk}

\author[mymainaddress]{Peng-Fei~Xian}
\ead[url]{xian.pf@my.cityu.edu.hk}

\author[mymainaddress]{Wei-Feng Ou}
\ead[url]{weifengou2-c@my.cityu.edu.hk}

\address[mymainaddress]{Department of Electrical Engineering, City University of Hong Kong, Hong Kong, China}

\begin{abstract}
    Real-time semantic segmentation has received considerable attention due to growing demands in many practical applications, such as autonomous vehicles, robotics, etc. Existing real-time segmentation approaches often utilize feature fusion to improve segmentation accuracy. However, they fail to fully consider the feature information at different resolutions and the receptive fields of the networks are relatively limited, thereby compromising the performance. To tackle this problem, we propose a light Cascaded Selective Resolution Network (CSRNet) to improve the performance of real-time segmentation through multiple context information embedding and enhanced feature aggregation. The proposed network builds a three-stage segmentation system, which integrates feature information from low resolution to high resolution and achieves feature refinement progressively. CSRNet contains two critical modules: the Shorted Pyramid Fusion Module (SPFM) and the Selective Resolution Module (SRM). The SPFM is a computationally efficient module to incorporate the global context information and significantly enlarge the receptive field at each stage. The SRM is designed to fuse multi-resolution feature maps with various receptive fields, which assigns soft channel attentions across the feature maps and helps to remedy the problem caused by multi-scale objects. Comprehensive experiments on two well-known datasets demonstrate that the proposed CSRNet outperforms the main-stream efficient semantic segmentation approaches by accuracy and can be performed in real-time reference. 
\end{abstract}

\begin{keyword}
    \textit{Semantic segmentation; Attention mechanism; Real-time inference; Deep neural networks}.
\end{keyword}

\end{frontmatter}



\section{Introduction}
\label{sec:introduction}
Semantic segmentation is a process of assigning dense semantic labels to each pixel in an image. Currently, semantic segmentation has gained remarkable improvement with deep learning, especially convolutional neural networks (CNN). Researchers are continually working on improving the segmentation accuracy and advance state-of-the-art performances \citep{yuan2019object, wang2020deep, tao2020hierarchical, yu2020context, zhang2020resnest, fu2020scene}. General semantic segmentation networks care less about inference speed and computation cost. Hence, they usually have large-scale backbones such as HRNet \citep{wang2020deep}, ResNet-101 \citep{he2016deep}, and Xception \citep{chollet2017xception}. With complex structures and deep layers, the network is able to extract comprehensive features from larger spatial ranges and benefit feature representation. However, heavy backbones often have high computational complexity and low inference speed. In recent years, real-time semantic segmentation has received considerable attention due to growing demands in autonomous vehicles, robotics, etc. These applications put a strict requirement on inference efficiency with low latency to make real-time implementations ($\geq 30$ Frames Per Second) and expect competitive segmentation performance, which is not suitable to employ heavy backbones. Therefore, how to balance efficient inference speed and high segmentation accuracy becomes a challenging task. 

Existing real-time semantic segmentation methods employ three strategies to accelerate the whole model: (1) Input resolution restriction. A down-sampling ratio $\gamma$ (usually 0.5 or 0.75) is utilized for the original image to obtain a smaller input resolution. Under the same network architecture, smaller inputs will consume fewer computing resources, resulting in higher inference speed. However, the down-sampling process may interfere with the detection of small objects and lose spatial information around boundaries, which degrade segmentation performance. (2) Channel pruning. As the backbone network goes from shallow to deep, the resolution of feature maps is getting smaller and smaller, and the number of feature channels is increasing. Pruning the channel numbers can decrease computational cost and improve inference speed, yet it will weaken feature extraction ability. (3) Shallow model architectures. Real-time semantic segmentation methods focus on achieving a decent trade-off between efficiency and accuracy, which means the backbone network should be a lightweight structure. The main drawback of lightweight architectures is that the extracted features cannot achieve sufficient receptive fields. Generally, current approaches leverage fusion of the above strategy to increase inference speed for real-time semantic segmentation.

\begin{figure}[t]
\centering
{\includegraphics[width=\textwidth]{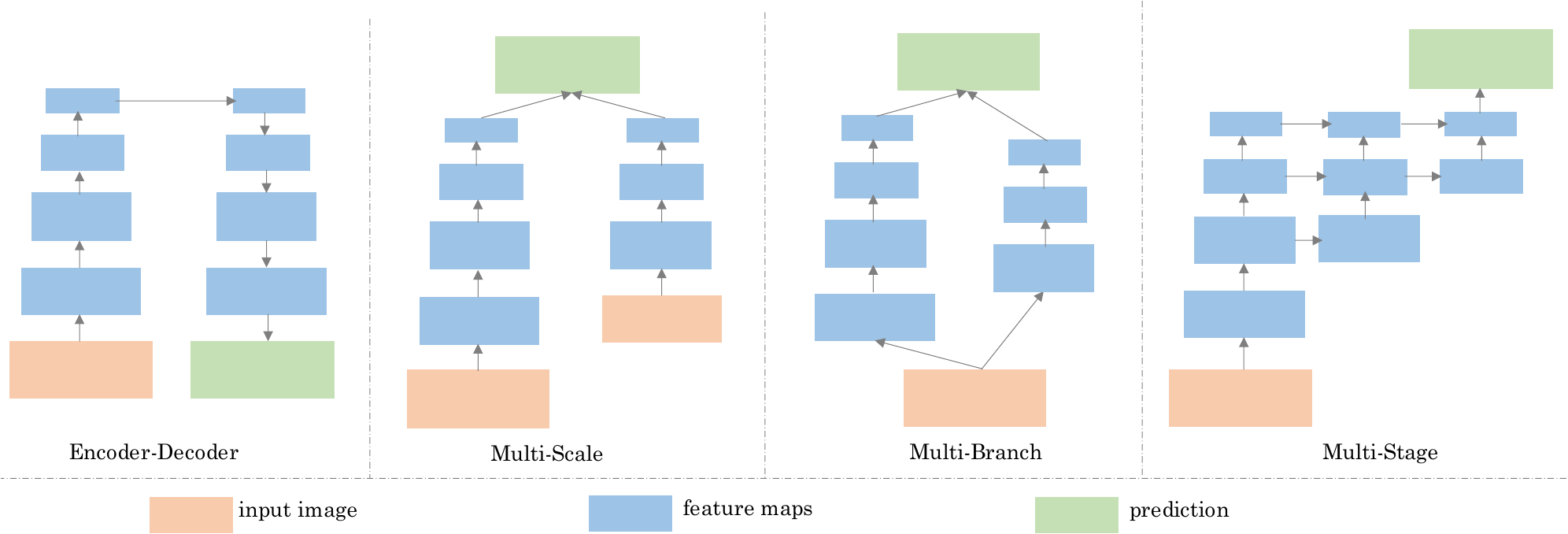}}
\caption[width=\linewidth]{Typical network structures for real-time semantic segmentation. From left to right: encoder-decoder network, multi-scale network, multi-branch network, multi-stage network. }
\label{fig:structure}
\end{figure}

In addition to inference speed, segmentation accuracy is another criterion to measure the quality of real-time semantic segmentation algorithms. Designing an organized network structure is one of the keys to achieving high segmentation accuracy. Popular real-time segmentation structures include encoder-decoder networks \citep{paszke2016enet,badrinarayanan2017segnet,orsic2019defense,hu2020real,arani2021rgpnet}, multi-scale networks \citep{zhao2018icnet,mazzini2018guided,orsic2019defense,yu2020context,li2021real}, multi-branch networks \citep{yu2018bisenet,poudel2019fast,yu2020bisenet}, and multi-stage networks \citep{li2019dfanet,zhuang2019shelfnet}, which are visualized in Figure \ref{fig:structure}. Note that some networks may develop a combination of the above structures. Encoder-decoder networks have skip connections to recover high-resolution features from low-resolution features. Multi-scale networks take advantage of multi-resolution inputs to extract more spatial features from larger scale input and collect more semantic features from smaller scale input. Apart from multi-scale networks, multi-branch networks design multiple branches to achieve this goal, where a deeper branch is employed to capture the global context and a shallow branch is developed to learn spatial details. Lastly, multi-stage networks aggregate discriminative features by cascading structures and implement skip-connection to recover image details in the decoder module. The network architecture proposed in this paper is a multi-stage network. With the help of the multi-stage structure, our network can strengthen the feature representation ability progressively. 

\begin{figure}[t]
\centering
{\includegraphics[width=\textwidth]{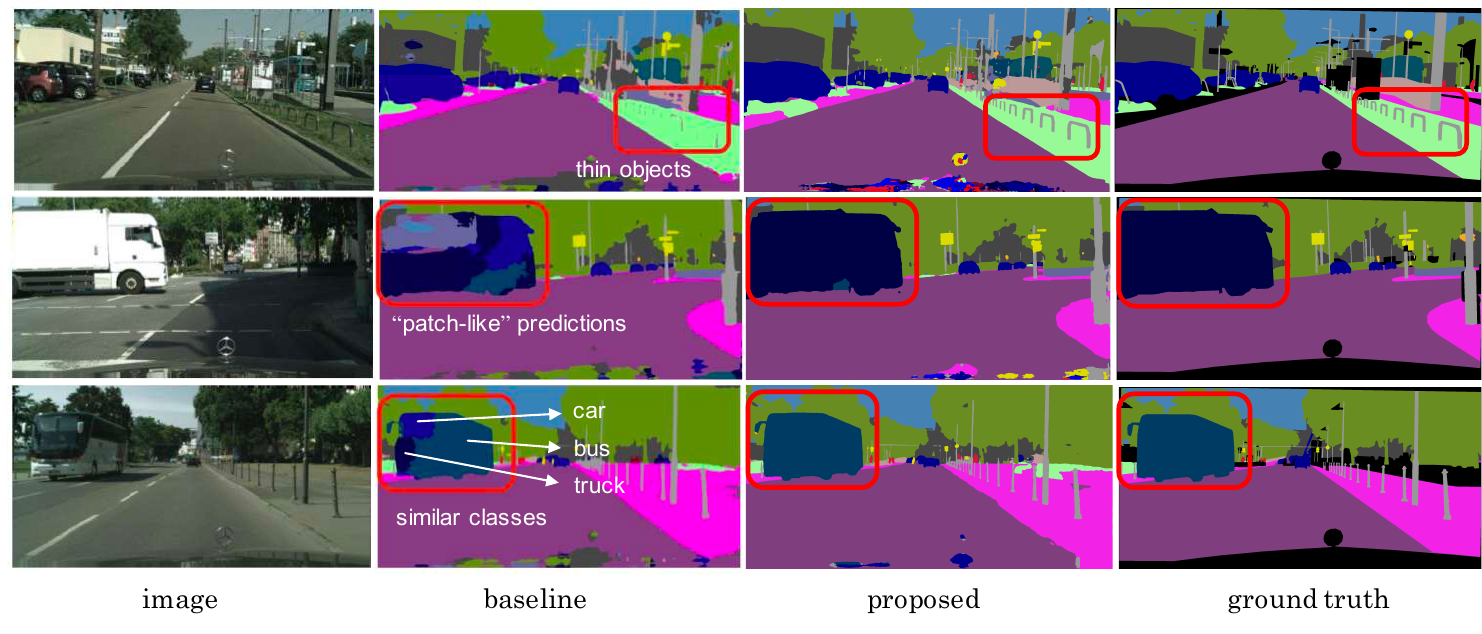}}
\caption[width=\linewidth]{Typical weaknesses for real-time semantic segmentation approaches: thin objects segmentation, dominant objects segmentation and similar classes segmentation.}
\label{fig:prob}
\end{figure}

Since real-time semantic segmentation approaches exploit lightweight networks which have limited layers and insufficient receptive fields, it mainly faces three problems. First, the downsampling operation leads to poor performances in segmenting some small structures like thin “poles”, small “traffic lights”, and “traffic signs”. An example of segmenting thin poles is shown in the top row of Figure \ref{fig:prob}. Second, there may exist “patch-like” discontinuous predictions in the interior of the dominant objects (e.g., nearby buses, trucks, or trains close to the camera). The category of the dominant objects may be misclassified into other classes, as shown in the middle row of Figure \ref{fig:prob}. Third, due to the shallow network, the feature extraction ability is weaker than that of the deep structure without a doubt. Some classes with similar shape and appearance are confusing in classification for a neural network, e.g., “bus” and “truck”. Therefore, it is easy to misclassify the “bus” pixel into the “truck” pixel and vice versa, as shown in the bottom row of Figure \ref{fig:prob}. More details can be found in Section 4.5. 

To tackle these issues and improve the performance for real-time segmentation, we propose a Cascaded Selective Resolution Network (CSRNet) by using progressive aggregation of features at different resolutions. In the feature aggregation process, CSRNet continues to incorporate global semantics and preserve edge details, thereby alleviating the problems of segmenting thin objects and similar classes. Specifically, the proposed CSRNet consists of two critical modules: (1) The Shorted Pyramid Fusion Module (SPFM) is employed to embed the global context information and enlarge the receptive field in each stage at the corresponding resolution of feature maps. The multiple adoption of SPFM enables rich context information embedding in the proposed network. (2) The Selective Resolution Module (SRM) based on the attention mechanism, is designed to fuse feature information of adjacent resolutions. SRM assigns soft channel attentions across the feature maps to handle the problem caused by multi-scale objects. More attention will be assigned to the lower-resolution feature map due to its sizeable receptive field if there is a dominant object inside the input image. A comparison of speed-accuracy trade-off with the previous real-time methods on testing set of Cityscapes \citep{cordts2016cityscapes} and Camvid \citep{brostow2008segmentation} is shown in Figure \ref{fig:comp}. The proposed method outperforms the previous efficient segmentation approaches by accuracy and can be run in real-time. The main contributions of our work are as follows:

\begin{figure}[t]
\centering
{\includegraphics[width=\textwidth]{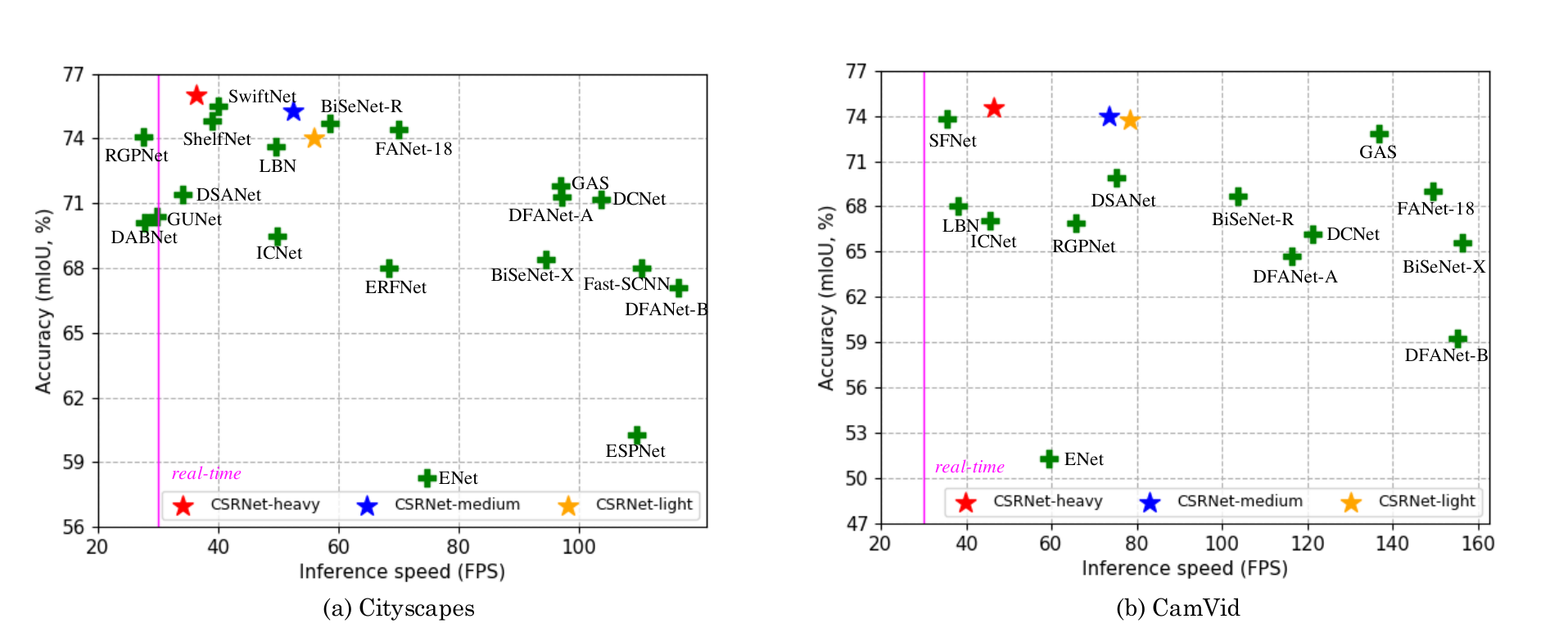}}
\caption[width=\linewidth]{A comparison of speed-accuracy trade-off on Cityscapes and Camvid testing set. The pentagrams indicate our methods while green crosses indicate other methods. Methods on the right side of the purple vertical line represent real-time implementations. More details can be found in Table \ref{table:citys} and Table \ref{table:camvid}.}
\label{fig:comp}
\end{figure}

\begin{itemize} 
\item \textbf{Multiple-stage segmentation network.}  We propose a multiple-stage segmentation network CSRNet to refine the feature maps progressively and improve the feature representation ability.

\item \textbf{Selective resolution module.}  We propose a novel selective resolution module (SRM), which exploits the attention mechanism to better aggregate multi-resolution feature maps with various receptive fields.

\item \textbf{Remedy the problem caused by multi-scale objects.} The proposed network can alleviate the problem caused by multi-scale objects, showing the superiority in segmenting the dominant objects.

\item \textbf{Experiments for real-time semantic segmentation.} The proposed network strikes a balance between segmentation accuracy and inference speed on two public challenging datasets: Cityscapes and CamVid.
\end{itemize}

\section{Related work}\label{sec:related}
This section will introduce the works related to our approach, such as real-time semantic segmentation, self-attention mechanism, and context information embedding. 

\subsection{Real-time Semantic Segmentation}
Compared with the general semantic segmentation tasks, the main challenge of real-time semantic segmentation is to strike a balance between segmentation accuracy and inference speed. Popular approaches like input resolution restriction, channel pruning, or shallow model architectures can accelerate the model by reducing the computation burden. However, all of these strategies sacrifice the segmentation accuracy to some extent, and the loss of spatial details caused by decreasing the input resolution is irreparable. The most typical way to remedy the decline of accuracy is to insert feature aggregation modules. Shallow layers in the neural networks tend to learn rich spatial and edge features from the local area, while deep layers can learn significant abstract features like context information. Designing a feature fusion module is an intuitive solution to fuse the features from different layers and combine various feature information for better feature representation capabilities \citep{zhao2018icnet,yu2018bisenet,li2019dfanet,zhuang2019shelfnet,arani2021rgpnet}. DFANet \citep{li2019dfanet} introduces the sub-network aggregation and sub-stage aggregation to achieve multi-scale feature propagation. BiSeNet \citep{yu2018bisenet} extracts spatial features and categorical semantic features separately via two independent paths and develops a feature fusion module to fuse features from different paths. 

Some works adopt multi-scale input to enhance segmentation quality. ICNet \citep{zhao2018icnet} takes cascaded image inputs (i.e., low-, medium-, and high-resolution images), which exploits the efficiency of processing low-resolution images vis a full segmentation backbone. Meanwhile, it can achieve high inference speed by feeding medium and high-resolution images into extremely light backbones. SwiftNet \citep{orsic2019defense} employs an image pyramid as input to improve segmentation accuracy. The lowest resolution of the image pyramid can enlarge the receptive field and regularize the encoder of the proposed model. Apart from multi-scale input, designing a proper loss function is another way to promote training performance. SwiftNet V2 \citep{orvsic2021efficient} increases the penalty for boundary pixels by proposing the boundary-aware loss to emphasize pixels near semantic boundaries. BiSeNet V2 \citep{yu2020bisenet} inserts the auxiliary segmentation head to different semantic branches and trains the network with the auxiliary loss.

\subsection{Self-attention Mechanism}
The self-attention mechanism was first adopted to show the global dependencies of different words in a specific sentence in natural language processing (NLP) \citep{bahdanau2014neural,luong2015effective,vaswani2017attention}. In segmentation task, the attention module, which is generally adopted on the higher layer of a neural network, enables the network to perform better by refining feature maps via the feature denoising process \citep{xie2019feature}. However, the traditional non-local block is unsuitable for real-time semantic segmentation due to its heavy computation cost when processing a high-resolution feature map. Many later methods have been proposed to reduce the computation burden of non-local blocks \citep{chen20182,yuan2019object,huang2019ccnet,li2019expectation,he2019adaptive,huang2019interlaced,cao2019gcnet}. The authors in \citep{huang2019ccnet,he2019adaptive,huang2019interlaced,cao2019gcnet} implemented consecutive sparse attention to replace the dense attention in the non-local networks, while the authors in \citep{chen20182,yuan2019object,li2019expectation} employed compact descriptors to reduce the computational complexity. In CCNet \citep{huang2019ccnet}, a criss-cross attention module was designed to generate a sparse attention map and exploits the recurrent criss-cross attention module to obtain richer and denser context in- formation. APCNet \citep{he2019adaptive} constructs multiple adaptive context modules, and each module transforms the feature map into a set of subregion representations. The context vector for each local position is calculated by leveraging the local affinity coefficients in each subregion. A2Net \citep{chen20182} first gathers key features from the entire feature space into global descriptors and adaptively distributes them to each position. OCRNet \citep{yuan2019object} obtains the object region representation and then computes the relation between each pixel position and each object region.

In addition to the methods that model the relationships among positions at different spatial locations, channel attention performs dynamic channel-wise feature recalibration by modeling interdependencies between different channels. Channel attention is more computation-friendly than spatial attention. The spatial attention assigns the same attention weights on the channel dimension, while the channel attention assigns the same attention weights on the spatial dimension \citep{hu2018squeeze,li2019selective,wang2020eca}. SENet \citep{hu2018squeeze}exploits global average pooling to generate channel-wise statistics and employs a sigmoid activation as a gating operation to determine the feature importance. Based on SENet, the authors in \citep{wang2020eca} proposed ECA-Net, which avoids dimensionality reduction in the SE module and develops a local cross-channel interaction strategy to guarantee efficiency. SKNet \citep{li2019selective} is inspired by SENet and Inception \citep{szegedy2015going} as a dynamic selection mechanism of receptive field size. In SK convolutions, multiple features are calculated from a single feature map with various kernel sizes to obtain different receptive fields. Soft attention across different channels is computed to select the generated multiple features adaptively. Besides, some works \citep{woo2018cbam,fu2019dual} combine spatial attention and channel attention simultaneously to achieve better segmentation performances.

\subsection{Context Information Embedding}
Convolutional neural networks are inherently limited to local receptive fields due to the fixed geometric structures. Two common ways to enlarge the receptive field include global average pooling and dilated convolution \citep{treml2016speeding,chen2017rethinking,chen2017deeplab,mehta2018espnet,mehta2019espnetv2,dong2020real}. DFNet \citep{yu2018learning} introduces the global context with global average pooling at the end of the backbone network. DeepLab \citep{chen2017deeplab} employs dilated convolution to control the receptive field by adjusting the rate. Besides, to mitigate the inconsistency problem of segmenting the objects with multiple scales inside a single image, constructing a feature pyramid to capture diverse context information is an effective way. Several works have further integrated feature pyramid and multi-scale feature ensemble, such as pyramid scene parsing (PSP) \citep{zhao2017pyramid} and atrous spatial pyramid pooling (ASPP) \citep{chen2017rethinking,chen2017deeplab}. PSPNet \citep{zhao2017pyramid} explores important global contextual information along with sub-region context, which fuses features under several pyramid scales by employing different scales of average pooling layers. ASPP generates multiple paths with various dilated convolutions to extract rich context information. DeepLab V3 \citep{chen2017rethinking} incorporates the global context information by adding the image-level features through a global average pooling to the ASPP module.

\section{Methodology} \label{sec:method}
In this section, we will introduce the backbone network and the whole architecture of CSRNet, and then present two key modules in the proposed CSRNet: the SPFM and the SRM. Finally, we will explain how to obtain the medium and light network versions from the original network architecture. 

\subsection{Backbone Network}
ResNet-18 \citep{he2016deep}, light Xception \citep{chollet2017xception}, MobileNet \citep{howard2017mobilenets,sandler2018mobilenetv2,howard2019searching} are the popular lightweight backbones in the real-time semantic segmentation. Researchers may also create a moderate depth structure that has low complexity and promote efficient training from scratch \citep{romera2017erfnet,poudel2019fast,li2019dabnet,lin2020graph,elhassan2021dsanet}. Xception and MobileNet series utilize depth-wise separable convolutions to reduce network parameters. From the computational cost perspective, the 3 $\times$ 3 depth separable convolution is computationally 8 to 9 times less than the conventional 3 $\times$ 3 convolution. MobileNet V2 \citep{sandler2018mobilenetv2} introduces two blocks to improve the performance: Linear Bottleneck and Inverted Residual Blocks. MobileNet series is particularly suitable for embedded device designs like mobile applications. However, depth-wise separable convolutions are not directly supported in GPU firmware (the cuDNN library), and the latest cuDNN makes special optimizations for standard 3 $\times$ 3 convolution. Therefore, MobileNetV2 tends to be slower than ResNet-18 on GPUs' performance. Consequently, we choose ResNet-18 as the backbone of our network since we run on GPU platform.

\begin{figure}[t]
\centering
{\includegraphics[width=\textwidth]{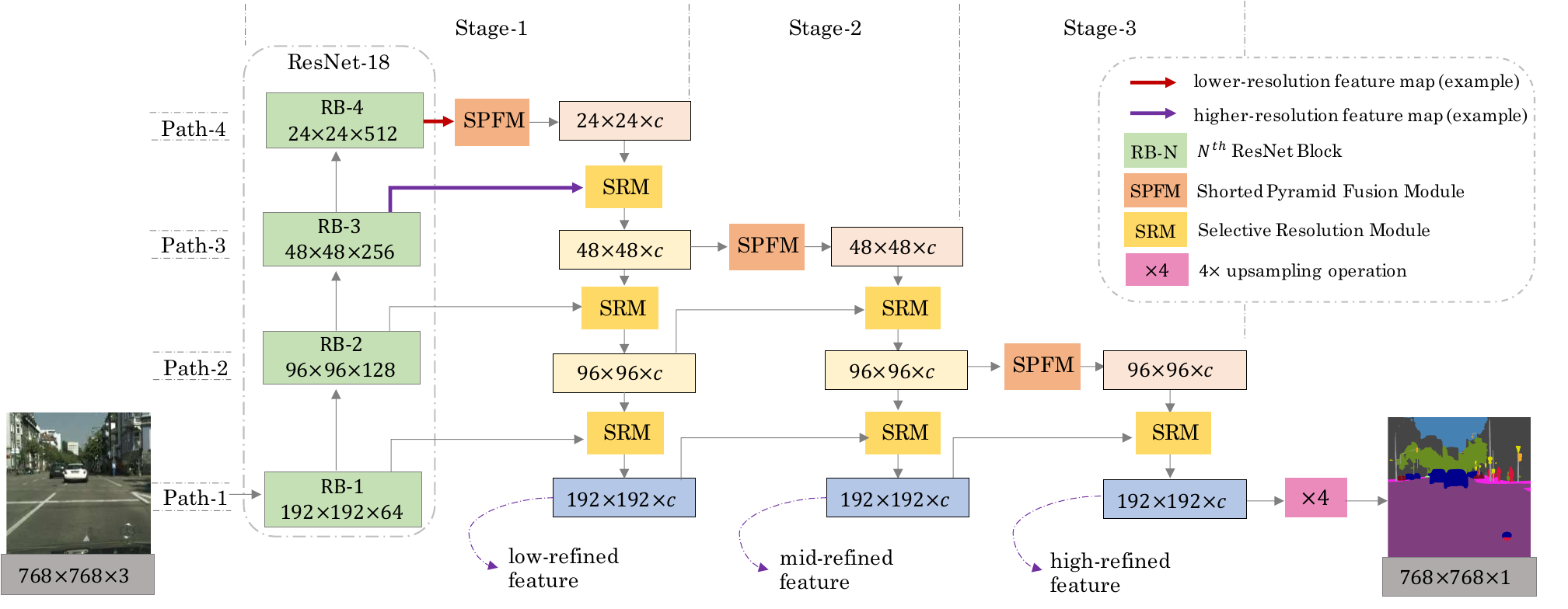}}
\caption[width=\linewidth]{An overview of the architecture of the proposed CSRNet. It contains 4 paths in the horizontal direction and 3 stages in the vertical direction. }
\label{fig:archi}
\end{figure}

\subsection{Network Architecture}
The proposed network has 4 paths in the horizontal direction and 3 stages in the vertical direction, as shown in Figure \ref{fig:archi}. The resolutions of the feature maps in each path are the same. Apart from the backbone, each stage in this structure aggregates the feature information from the lower-resolution feature map to the higher-resolution feature map through fusing features in adjacent paths. Stage-1 has a U-shape structure with a ResNet-18 backbone. Given an input $\mathbf{X} \in \mathbb{R}^{H \times W \times N} $ where it is normally a RGB image consisting of dimension height $H$, width $W$ and channel $N=3$, the spatial resolutions of four feature maps generated by ResNet Blocks (RB-1 to RB-4) are 4 $\times$, 8 $\times$, 16 $\times$, 32 $\times$ down-sampling of the input spatial resolution, and the channel numbers increase to 64, 128, 256 and 512, respectively. For real-time inference, the channel pruning is employed to reduce the channel numbers of the first three features in Stage-1. The number of output channels becomes 128 after the channel pruning process. The 32 $\times$ feature map outputted by RB-4 is fed into the SPFM to extract global context information. This process helps refine the spatial information by adding the global context to the corresponding feature map.

Next, the feature maps in adjacent paths are aggregated through the SRM. The SRM takes the lower-resolution feature map and the higher-resolution feature map as input and assigns soft attentions across feature channels. Three consecutive SRMs combine the high-level features and low-level details and output a 4 $\times$ down-sampling feature map, called the low-refined feature map. The process of Stage-2 and Stage-3 are similar to Stage-1, but we feed 16 $\times$ and 8 $\times$ feature maps outputted by previous stages into the SPFM, respectively. In this way, we can obtain the mid-refined feature map and high-refined feature map. The multi-stage process can be regarded as a feature refinement process, where the feature outputted by each stage becomes more and more representative and determinate. We will show the effectiveness of feature refinement in Section \ref{sec:exp}. Finally, the high-refined feature map goes through a 1 $\times$ 1 convolutional layer and a SoftMax operation to generate the segmentation prediction. This prediction is unsampled by a factor of 4 to obtain the final segmentation map.

\subsection{Shorted Pyramid Fusion Module (SPFM)}
The SPFM is designed to efficiently enlarge the receptive field at each stage. SPFM operates on the high-level feature map in the corresponding path at each stage, providing effective global contextual prior. In Stage-1, Stage-2, and Stage-3, SPFM is applied on 32 $\times$, 16 $\times$, 8 $\times$ down-sampling feature maps, respectively. The structure of SPFM is similar to Pyramid Pooling Module (PPM) in PSPNet \citep{zhao2017pyramid}, and we modify the architecture of PPM to reduce the computational burden and meet the requirements of real-time inference. Before the pyramid pooling operation with several different pyramid scales, the feature map (e.g., size: $ a \times b \times c_{o}$) is passed to a 1 $\times$ 1 convolutional layer followed by batch normalization and ReLU activation function to transform the dimensionality to a fixed smaller number (size: $ a \times b \times c$, $c \leq c_{o}$). $c = 128$ is a typical setting in our experiments. Then three feature maps with varying levels of details are produced by pooling operations over coarse spatial grids 2 $\times$ 2, 4 $\times$ 4, and 8 $\times$ 8, following by convolution, upsampling and concatenation layers to form the feature representation (size: $ a \times b \times 2c$). Finally, the feature representation is fed into a 1 $\times$ 1 convolutional layer for channel reduction and obtain the final feature map (size: $ a \times b \times c$). Compared with the PSPNet that operates PPM on the feature map of the last convolutional layer and receives the prediction directly, our SPFM works on three feature maps of different resolutions in the corresponding path. The global contextual priors of Stage-1 to Stage-3 are from different resolutions of feature maps. The low-refined feature map contains global information from the 32 $\times$ feature map. Likewise, the mid-refined feature map and the high-refined feature map contain the global information from the 16 $\times$, 8 $\times$ feature map, respectively. Finally, the cascaded structure enables the network to propagate global information in different stages and embed more comprehensive context information.

\begin{figure}[t]
\centering
{\includegraphics[width=7cm]{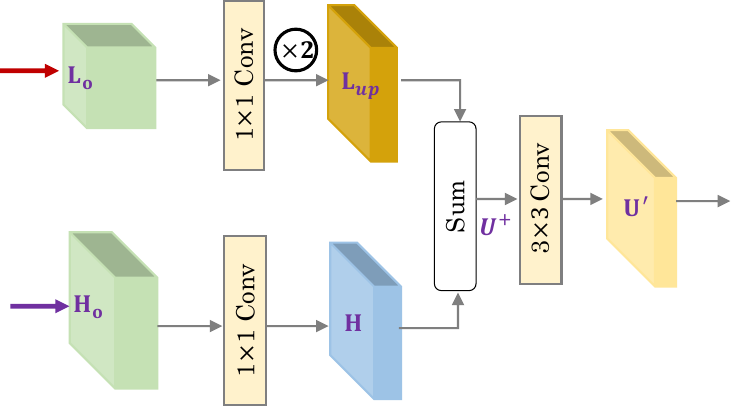}}
\caption[width=\linewidth]{An example of a simple way to fuse the information of different feature maps with diverse spatial resolutions. }
\label{fig:add}
\end{figure}

\begin{figure}[t!]
\centering
{\includegraphics[width=\textwidth]{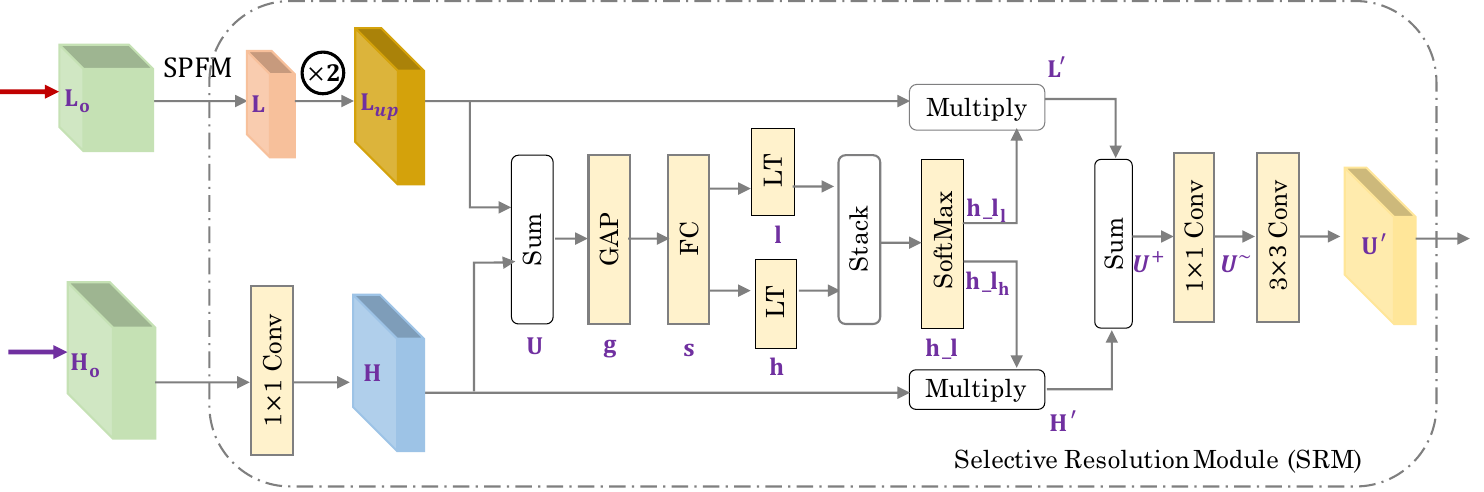}}
\caption[width=\linewidth]{The details of SRM, taking the fusing process of the outputs of RB-3 and RB-4 in Stage-1 as an example. The SRM integrates the attention mechanism and aggregates the feature maps with different resolution and receptive fields. SPFM: Shorted Pyramid Fusion Module. GAP: Global Average Pooling. $\times$ 2: 2 times upsampling using bilinear interpolation. FC: Fully Convolutional layer followed by a batch-normalization layer and a ReLU layer. LT: Linear Transformation. 1 $\times$ 1 or 3 $\times$ 3 Conv: 1 $\times$ 1 or 3 $\times$ 3 convolutional layer followed by a batch-normalization layer and a ReLU layer.}
\label{fig:srm}
\end{figure}

\subsection{Selective Resolution Module (SRM)}
Feature aggregation is widely adopted in current real-time semantic segmentation approaches, yet few researchers explore how to aggregate two specific feature maps with different resolutions. The simplest way to merge the information of different feature maps with various spatial resolutions is to bilinearly upsample the lower-resolution feature map to the higher-resolution maps and do the element-wise summation. This process can be formulated as:

\begin{equation}
\mathbf{U}^{+}=\mathbf{H}+\mathbf{L}_{up}.             
\end{equation}

If the channel numbers of two feature maps are not equal, a 1 $\times$ 1  convolution can be applied to match the channel number. After summation operation, a 1 $\times$ 1 or 3 $\times$ 3 convolution followed by batch normalization and ReLU activation is adopted to integrate the feature map further, as shown in Figure \ref{fig:add}. This method is extremely efficient since only an upsampling and several convolution operations are included in the feature aggregation process. However, merely adding two feature maps damages the receptive information contained in the original feature maps.

The original lower-resolution feature map tends to have a larger receptive field and can extract more contextual information. Moreover, the receptive field of the feature map after SPFM is significantly enlarged. The higher-resolution feature map has a shallow receptive field, but it can extract more low-level information such as edges and texture from the local area. Instead of directly adding the feature maps, can we construct a module to determine the importance of the feature maps with different receptive fields when fusing these feature maps? Attention is a possible answer. We can learn soft channel attentions across feature maps with different resolutions, i.e., if there is a dominant object inside the input image, more attention desires to be assigned to the lower-resolution feature map due to its sizeable receptive field and vice versa.

Inspired by SKNet \citep{li2019selective}, we propose a feature aggregation module SRM, which exploits the attention mechanism to better aggregate feature maps with various resolutions and receptive fields. The structure of SRM is shown in Figure \ref{fig:srm}, taking the fusing process of the outputs ($\mathbf{H}_{o}$ and $\mathbf{L}_{o}$) of RB-3 and RB-4 in Stage-1 as an example. Given an image input $\mathbf{X} \in \mathbb{R}^{H \times W \times 3} $, after 1 $\times$ 1 convolution and SPFM, $\mathbf{H} \in \mathbb{R}^{\frac{1}{16} H \times \frac{1}{16} W \times c }$ and $\mathbf{L} \in \mathbb{R}^{\frac{1}{32} H \times \frac{1}{32} W \times c }$ are fused through the SRM. In Figure \ref{fig:srm}, $H=W=768$ and $c=128$. First, $\mathbf{L}$ is upsampled to $\mathbf{L}_{up}=\mathcal{F}_{up}(\mathbf{L})$, where $=\mathcal{F}_{up}$ is the bilinear interpolation, to match the dimensional size of $\mathbf{H}$ for element-wise summation:

\begin{equation}
\mathbf{U}=\mathbf{H}+\mathbf{L}_{up}.             
\end{equation}

After combining the feature information with the simple summation, we seek dual-channel descriptors to capture channel-wise dependencies corresponding to the two feature maps. Therefore, we extract a global descriptor from the summed feature map and split it into two descriptors. We employ a global average pooling for each channel of $\mathbf{U}$ independently to squeeze the global spatial information and obtain a descriptor $\mathbf{g} \in \mathbb{R}^{c} $. Formally, $\mathbf{g}$ is generated by averaging $\mathbf{U}$ through spatial dimensions $h \times w$, where $h=\frac{1}{16} H $ and $w=\frac{1}{16} W $. The $t^{th}$ element of $\mathbf{g}$ can be calculated by:

\begin{equation}
\mathbf{g}_{t}=\mathcal{F}_{gap} (\mathbf{U}_{t} )=\frac{1}{h \times w} \ \sum_{i=1}^{h}\ \sum_{j=1}^{w} \mathbf{U}_{t}(i,j). 
\end{equation}

To ensure the capability of the global descriptor, we employ a fully connected (FC) layer without channel dimensionality reduction to learn a more effective descriptor $\mathbf{s}$:

\begin{equation}
\mathbf{s}=\mathcal{F}_{fc}(\mathbf{g})=\delta(\mathcal{B}(\mathcal{W}_{1}\mathbf{g})), 
\end{equation}
where $\delta$ is the ReLU non-linearity, $\mathcal{B}$ is the Batch Normalization \citep{ioffe2015batch}, $\mathcal{W}_{1}  \in \mathbb{R}^{c \times c}$ contains the parameters of the fully connected layer. Then the channel descriptors $\mathbf{h}$ and $\mathbf{l}$ are obtained by applying two linear transformations to the incoming $\mathbf{s}$. Afterwards, to normalize $\mathbf{h}$ and $\mathbf{l}$ across the channel dimension, $\mathbf{h}$ and $\mathbf{l}$ are stacked, and a $SoftMax$ operator is adopted to generate soft attention across channels $\mathbf{h}\_\mathbf{l}$:

\begin{equation}
\mathbf{h}=\mathcal{W}_{2}\mathbf{s} + b_{2},
\mathbf{l}=\mathcal{W}_{3}\mathbf{s} + b_{3},
\end{equation}

\begin{equation}
\mathbf{h}\_\mathbf{l}=\sigma(\mathcal{F}_{stack}(\mathbf{h},\mathbf{l})),
\end{equation}
where $\mathcal{W}_{2}  \in \mathbb{R}^{c \times c}$, $\mathcal{W}_{3}  \in \mathbb{R}^{c \times c}$, $b_{2}  \in \mathbb{R}^{c}$, $b_{3}  \in \mathbb{R}^{c}$ are the weights and bias of the two linear transformations. $\sigma$ denotes SoftMax operation. We separate two normalized soft attention vectors $\mathbf{h}\_\mathbf{l}_{h}$ and $\mathbf{h}\_\mathbf{l}_{l}$ from $\mathbf{h}\_\mathbf{l}$. Next, the values of the two descriptors are multiplied to the higher-resolution feature map $\mathbf{H}$ and lower-resolution feature map $\mathbf{L}$. Since the higher-resolution feature map and the lower-resolution feature map are separately outputted from different residual blocks, a simple weighted summation can discriminate the importance of the feature maps with various receptive fields when fusing, but it cannot integrate the information insides the feature maps effectively. To tackle this problem, two convolutional layers are employed to further aggregate the channel and spatial information of the merged feature map at the end of SRM. A 1 $\times$ 1 convolutional layer followed by a batch-normalization layer and a ReLU layer is utilized to aggregate information in channel dimension, and a 3 $\times$ 3 convolution followed by a batch-normalization layer and a ReLU layer further blends the features.

\begin{equation}
\mathbf{H}^{'}=\mathbf{h}\_\mathbf{l}_{h}\cdot\mathbf{H},\mathbf{L}^{'}=\mathbf{h}\_\mathbf{l}_{l}\cdot\mathbf{L}_{up},
\end{equation}

\begin{equation}
\mathbf{U}^{+}=\mathbf{H}^{'}+\mathbf{L}^{'}=\mathbf{h}\_\mathbf{l}_{h}\cdot\mathbf{H}+\mathbf{h}\_\mathbf{l}_{l}\cdot\mathbf{L}_{up},
\end{equation}

\begin{equation}
\mathbf{U}^{\sim}=\delta(\mathcal{B}(\mathcal{W}_{4}(\mathbf{H}^{'}+\mathbf{L}^{'}))), 
\end{equation}

\begin{equation}
\mathbf{U}^{'}=\delta(\mathcal{B}(\mathcal{W}_{5}\mathbf{U}^{\sim})), 
\end{equation}

where $\mathbf{h}\_\mathbf{l}_{h}$ and $\mathbf{h}\_\mathbf{l}_{l}$ are from $\mathbf{h}\_\mathbf{l}$, denoting the soft attention vector for $\mathbf{H}$ and $\mathbf{L}$. $\mathcal{W}_{4}  \in \mathbb{R}^{1 \times 1 \times c \times c}$ and $\mathcal{W}_{5}  \in \mathbb{R}^{3 \times 3 \times c \times c}$ are the weights for the 1 $\times$ 1 convolutional layer and 3 $\times$ 3 convolutional layer, respectively.

The main differences between SRM and Selective Kernel (SK) Unit lie in three aspects. (1) The input of an SK unit is a single feature map, followed by two convolutions with varying sizes of kernels to generate two feature maps with diverse receptive fields. In contrast, the inputs of SRM are two different feature maps which have various receptive fields already. Moreover, the receptive field of the feature map in the lower-resolution path after SPFM is significantly enlarged. The difference of receptive field of SRM inputs is larger than that of SK Unit inputs. (2) In the SK Unit, a compact feature is created from the global descriptor by a simple FC layer, with the shrinking of channel number for better efficiency. SRM does not reduce the channel number due to the observation found by \citep{wang2020eca} that dimensionality reduction destroys the direct correspondence between the channel and its weight, while avoiding dimensionality reduction helps learn more effective channel attention. Besides, since we have applied channel reduction in previous steps, the dimensionality $ c $ in the FC layer is small. Keeping dimensionality unchanged still has a low computational capacity. (3) Both the SK module and SRM have two convolutional layers, but the intentions of convolution are absolutely different. The two convolutional operations at the beginning of SK module aim to produce two feature maps with different kernel sizes, while the two convolutional operations at the end of SRM further aggregate the feature information of the merged feature map. The convolution is significant to SRM, the effectiveness of which will be validated in the ablation study.

\begin{figure}[t]
\centering
{\includegraphics[width=\textwidth]{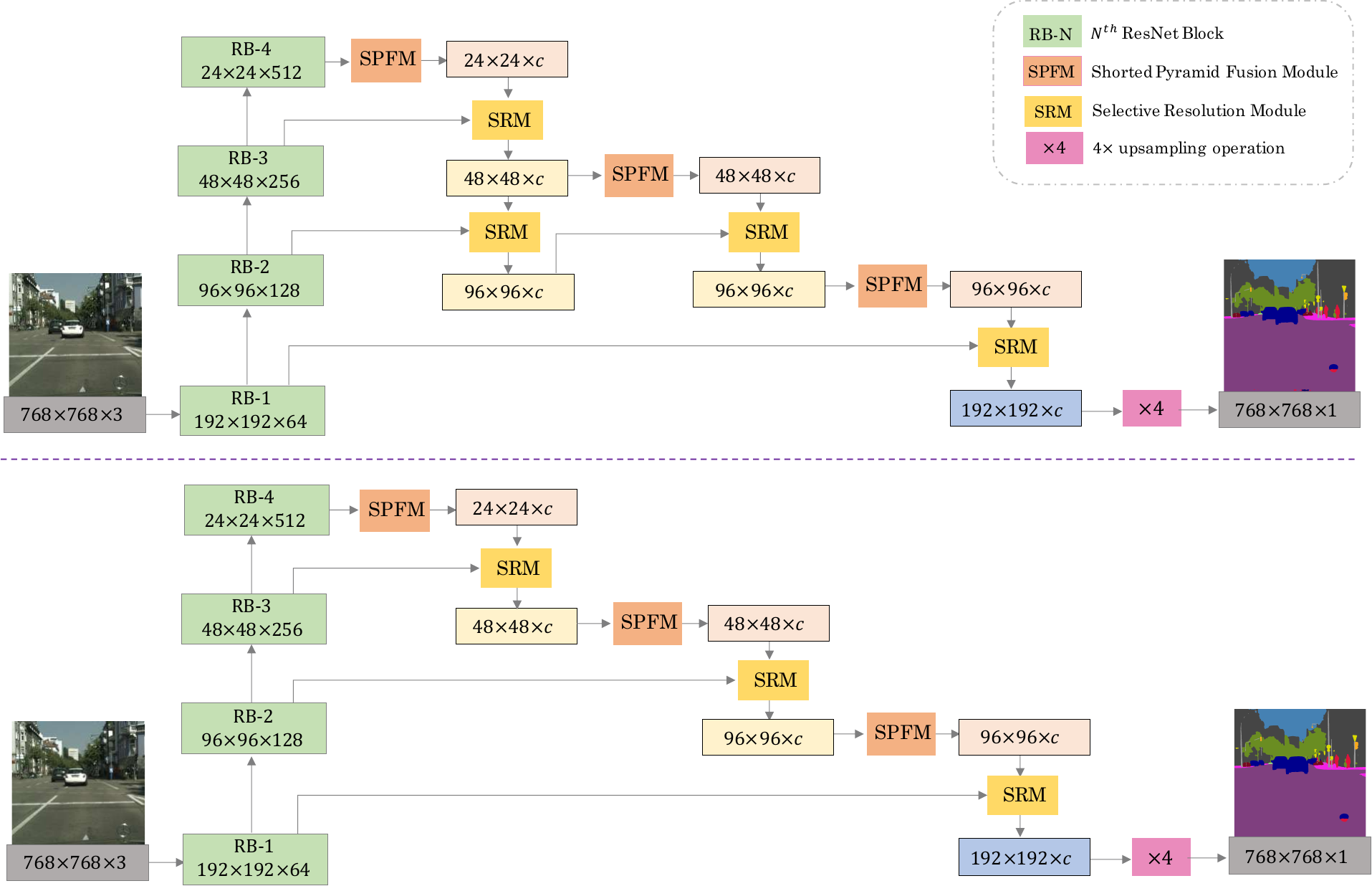}}
\caption[width=\linewidth]{The architecture of medium and light network. The upper image shows the architecture of CSRNet-medium and the bottom image shows the architecture of CSRNet-light. }
\label{fig:three}
\end{figure}

\subsection{Medium and Light Network}
From the network architecture, it is evident that Path-1 occupies a heavier computational capacity than Path-2, and Path-2 occupies a heavier computation capacity than Path-3, etc. Calculations on the high-resolution feature map usually require more computation resources than calculations on the low-resolution feature. The spatial input size is downsampled by factors of 4, 8, 16, and 32 in Path-1 to Path-4. Since the calculation on the feature map of 4 $\times$ down-sampling resolution is very time-consuming, we can remove the 1 $\times$ 1 convolution and SRM of Stage-1 and Stage-2 in Path-1 and only keep the last 1 $\times$ 1 convolutional layer and SRM. The simplified network is called CSRNet-medium (the upper image in Figure \ref{fig:three}). Similarly, the 1 $\times$ 1 convolution and SRM of Stage-1 in Path-2 can also be deleted, leading to the simplest network CSRNet-light (the bottom image in Figure \ref{fig:three}). The medium and light version of networks avoid heavy computation on large resolution feature maps to further boost inference speed. The original network is called CSRNet-heavy. Note that the all versions of the proposed network are capable of real-time semantic segmentation. CSRNet-heavy extracts the global information from high-level feature maps using SPFM, and aggregates the global semantics to the lowest-level feature map which contains rich spatial information. The multiple adoption of SPFM enable rich context information embedding in different stages and significantly enlarge the receptive field. Both CSRNet-medium and CSRNet-light reserve the design of SPFM in different paths and discard the SRM on larger resolution phases, which can decrease computational complexity and avoid dramatic drop in accuracy.

\section{Experiments and Results} \label{sec:exp}
In this section, we introduce the basic information of the datasets and the implementation details of CSRNet. Ablation studies of different components in three versions of the network are conducted. We also provide a comprehensive explanation of experimental results. For quantitative evaluation, the standard metric of the mean intersection over union (mIoU) is adopted as the criteria of segmentation accuracy, and the Time and Frames Per Second (FPS) are adopted as the criteria of inference speed.

\subsection{Datasets}
\textbf{Cityscapes Dataset:} The cityscapes dataset \citep{cordts2016cityscapes} focuses on the semantic understanding of urban street scenes. In this dataset, images were collected from a car perspective in 50 cities with good or medium weather conditions. This dataset includes 5000 images with fine annotations and 20,000 images with coarse annotations. We merely utilize the fine annotations to train and validate our model. The annotations contain 30 classes, 19 of which are chosen as the labels of semantic segmentation task. The resolution of images in this dataset is 1024 $\times$ 2048. We follow the standard setting to partition the images into three sets: training, validation, and testing with 2975, 500, and 1525 images, respectively.

\textbf{CamVid Dataset}: Cambridge-driving Labeled Video Database \citep{brostow2008segmentation} is an early road scene dataset captured from the perspective of a driving automobile, containing 701 images with resolution 960 $\times$ 720. The images and annotation quality of this database are lower than that of Cityscapes, which makes this dataset challenging. Images in this dataset are extracted from three different video sequences. We adopt the same setting in \citep{badrinarayanan2017segnet} by grouping the 32 original classes into 11 classes through combining the same category of classes. The images are split into training, validation, and testing set with 367, 101, and 233 images, respectively. Following the pioneering work \citep{orsic2019defense}, we train our model on the combined training and validation subsets and test the accuracy on the testing set.

\subsection{Implementation Details}
\textbf{Training.} We train all of our models using the Adam \citep{kingma2014adam} optimizer with the initial learning rate set to 4e-4 in Cityscapes and 8e-4 in CamVid. For all datasets, we adopt 8 batch size. We decay the learning rate with cosine annealing strategy \citep{ilya2016sgdr} to the minimum value of 1e-6 in the last epoch. The weight decay is set to 1e-4. We update the pre-trained ResNet-18 backbone parameters with 4 times smaller learning rate and apply 4 times smaller weight decay than the randomly initialized parameters. We employ random horizontal flip, random scale, and random crop the image to a fixed size for training. The random scales contain {0.75, 1.0, 1.25, 1.5, 1.75, 2.0}. We adopt the cropped resolution of 768 $\times$ 768 for Cityscapes and 960 $\times$ 720 for CamVid. We employ the conventional cross-entropy loss to train our network, with 250 epochs on Cityscapes and 400 epochs on CamVid. We do not adopt other evaluation tricks, like sliding-window evaluation and multi-scale testing.

\textbf{Setup.} We conduct all experiments based on PyTorch 1.1.0 framework with one NVIDIA GeForce GTX 1080 Ti under CUDA 10.2 and cuDNN 7.6.5.

\subsection{Performances}
This subsection compares our networks with other state-of-the-art methods on two benchmarks: Cityscapes and CamVid.

\subsubsection{Accuracy and Inference Speed Comparison}
We present the segmentation accuracy and inference speed of the proposed network on Cityscapes and CamVid dataset. Table \ref{table:citys} reports the comparison results of our method and state-of-the-art methods on the Cityscapes testing set. In this table, mIoU (val) and mIoU (test) indicate the mean IoU of Cityscapes validation set and testing set. The scaling factors of computing the normalized FPS are: 1 for GTX 1080 Ti, 0.44 for Tesla K40, 0.79 for Tesla K80, 0.61 For Titan X, 1.12 for Titan XP, and 1.37 for RTX 2080Ti. As shown in this table, our method achieves 75.3\% mean IoU with a speed of 52.5 FPS under CSRNet-medium network and yields 76.0\% mean IoU with 36.3 FPS under CSRNet-heavy network. Although CSRNet-heavy obtains a subtle higher mean IoU accuracy, it sacrifices nearly 16 FPS compared to the FPS achieved by CSRNet-medium. CSRNet-medium can run at high speed while accomplishing high-quality segmentation, which shows a better trade-off between efficiency and accuracy. Table \ref{table:camvid} reports the comparison results of our method and state-of-the-art method on the CamVid testing set. As shown in this table, our method reaches 73.9\% mean IoU with 73.5 FPS under CSRNet-medium network and delivers 74.5\% mean IoU with 46.4 FPS under CSRNet-heavy network. The segmentation accuracies on both datasets have achieved state-of-the-art performances and satisfied the real-time inference requirements.

\begin{table}[t]
	\renewcommand{\arraystretch}{0.9}
	\caption{Comparison of segmentation performance with several real-time state-of-the-art semantic segmentation methods on validation set and testing set of Cityscapes dataset. Table ordered by published time. “-” indicates that the corresponding result is not provided by the method.}
	\centering
	\label{table:citys}
	\resizebox{\columnwidth}{!}{
		\begin{tabular}{lccccccc}
        \hline
        \multirow{2}* {Method} & \multirow{2}* {GPU} & \multirow{2}* {Backbone} & mIoU  & mIoU  & Time & \multirow{2}* {FPS} & FPS\\
         & &          & (val)     & (test)      &   (ms)     &      & norm\\
        \hline
        ENet \citep{paszke2016enet} & Titan X & None & - & 58.3 & 13.0 & 76.9 & 74.7\\
        ERFNet \citep{romera2017erfnet} & Titan X &  None & 70.0 & 68.0 & 24.0 & 41.7 & 68.4\\
        ESPNet \citep{mehta2018espnet} & Titan X &  ESPNet & - & 60.3 & 8.9 & 112.9 & 109.6\\
        GUNet \citep{mazzini2018guided} & Titan Xp &  DRN-D-22 & 69.6 & 70.4 & 30.0 & 33.3 & 29.7\\
        BiSeNet-X \citep{yu2018bisenet} & Titan Xp &  Xception-39 & 69.0 & 68.4 & 9.5 & 105.8 & 94.5\\
        BiSeNet-R \citep{yu2018bisenet} & Titan Xp &  ResNet-18 & 74.8 & 74.7 & 15.3 & 65.5 & 58.5\\
        ICNet \citep{zhao2018icnet} & Titan X &  PSPNet-50 & - & 69.5 & 33 & 30.3 & 49.7\\
        ESPNet V2 \citep{mehta2019espnetv2} & Titan X &  ESPNet V2 & 66.4 & 66.2 & - & - & -\\
        Fast-SCNN \citep{poudel2019fast} & Titan Xp &  None & 68.6 & 68.0 & 8.1 & 123.5 & 110.3\\
        DABNet \citep{li2019dabnet} & GTX 1080Ti &  None & - & 70.1 & 36.1& 27.7 & 27.7\\
        DFANet-A \citep{li2019dfanet} & Titan X &  Xception A & - & 71.3 & 10 & 100 & 97.1\\
        DFANet-B \citep{li2019dfanet} & Titan X &  Xception B & - & 67.1 & 8.3 & 120 & 116.5\\
        SwiftNet \citep{orsic2019defense} & GTX 1080Ti &  ResNet-18 & 75.4 & 75.5 & 25.1& 39.9 & 39.9\\
        ShelfNet \citep{zhuang2019shelfnet} & GTX 1080Ti &  ResNet-18 & - & 74.8 & 25.6 & 39 & 39\\
        FANet-18 \citep{hu2020real} & Titan X &  ResNet-18 & 75.0 & 74.4 & 13.9 & 72 & 69.9\\
        LBN \citep{dong2020real} & Titan X &  LBN-AA & - & 73.6 & 19.6 & 51 & 49.5\\
        GAS \citep{lin2020graph} & Titan Xp &  GCN-Guided & - & 71.8 & 9.2 & 108.4 & 96.8\\
        RGPNet \citep{arani2021rgpnet} & GTX 2080Ti &  ResNet-18 & - & 74.1 & 26.5 & 37.8 & 27.6\\
        Lite-HRNet \citep{yu2021lite} & V100 &  Lite-HRNet-18 & 73.8 & 72.8 & - & - & -\\
        DCNet \citep{li2021real} & GTX 2080Ti &  ResNet-18 & - & 71.2 & 7.0 & 142 & 103.6\\
        DSANet \citep{elhassan2021dsanet} & GTX 1080Ti &  None & 79.8 & 71.4 & 29.3 & 34.1 & 34.1\\
        \hline
        CSRNet-light & GTX 1080Ti &  ResNet-18 & 76.1 & 74.0 & 17.9 & 56 & 56\\
        CSRNet-medium & GTX 1080Ti &  ResNet-18 & 76.6 & 75.3 & 19.0 & 52.5 & 52.5\\
        CSRNet-heavy & GTX 1080Ti &  ResNet-18 & 77.3 & 76.0 & 27.5 & 36.3 & 52.5\\
        \hline
		\end{tabular}
	}
\end{table}

\begin{table}[t]
	\renewcommand{\arraystretch}{0.9}
	\caption{Comparison of segmentation performance with several state-of-the-art real-time semantic segmentation methods on testing set of CamVid dataset. Table ordered by published time. “-” indicates that the corresponding result is not provided by the method.}
	\centering
	\label{table:camvid}
	\resizebox{\columnwidth}{!}{
		\begin{tabular}{lcccccc}
        \hline
        \multirow{2}* {Method} & \multirow{2}* {GPU} & \multirow{2}* {Backbone}  & mIoU  & Time & \multirow{2}* {FPS} & FPS\\
         & &            & (test)      &   (ms)     &      & norm\\
        \hline
        ENet \citep{paszke2016enet} & Titan X & None & 51.3 & 16.3 & 61.2 & 59.4\\
        ICNet \citep{zhao2018icnet} & Titan &  PSPNet-50 & 67.1 & 36.0 & 27.8 & 45.6\\
        BiSeNet-X \citep{yu2018bisenet} & Titan Xp &  Xception-39 & 65.6 & 5.7 & 175 & 156.3\\
        BiSeNet-R \citep{yu2018bisenet} & Titan Xp &  ResNet-18 & 68.7 & 8.6 & 116.3 & 103.8\\
        DFANet-A \citep{li2019dfanet} & Titan X & Xception A & 64.7 & 8.3 & 120 & 116.5\\
        DFANet-B \citep{li2019dfanet} & Titan X &  Xception B & 59.3 & 6.3 & 160 & 155.3\\
        SwiftNet \citep{orsic2019defense} & GTX 1080Ti &  ResNet-18 & 73.9 &- & - & -\\
        FANet-18 \citep{hu2020real} & Titan X &  ResNet-18 & 69 & 6.5 & 154 & 149.5\\
        LBN \citep{dong2020real} & Titan X &  LBN-AA & 68.0 & 25.4 & 39.3 & 38.2\\
        GAS \citep{lin2020graph} & Titan Xp &  GCN-Guided & 72.8 & 6.5 & 153.1 & 136.7\\
        SFNet \citep{li2020semantic} & GTX 1080Ti &  ResNet-18 & 73.8 & 28.2 & 35.5 & 35.5\\
        RGPNet \citep{arani2021rgpnet} & GTX 2080Ti &  ResNet-18 & 66.9 & 11.1 & 90.2 & 65.8\\
        DCNet \citep{li2021real} & GTX 2080Ti &  ResNet-18 & 66.2 & 6.0 & 166 & 121.2\\
        DSANet \citep{elhassan2021dsanet} & GTX 1080Ti &  None & 69.9 & 13.2 & 75.3 & 75.3\\
        \hline
        CSRNet-light & GTX 1080Ti &  ResNet-18 & 73.7 & 12.8 &  78.4 &  78.4\\
        CSRNet-medium & GTX 1080Ti &  ResNet-18 & 74.0 & 13.6 & 73.5 & 73.5\\
        CSRNet-heavy & GTX 1080Ti &  ResNet-18 & 74.5 & 21.6 & 46.4 & 46.4\\
        \hline
		\end{tabular}
	}
\end{table}

\begin{figure}[t]
\centering
{\includegraphics[width=\textwidth]{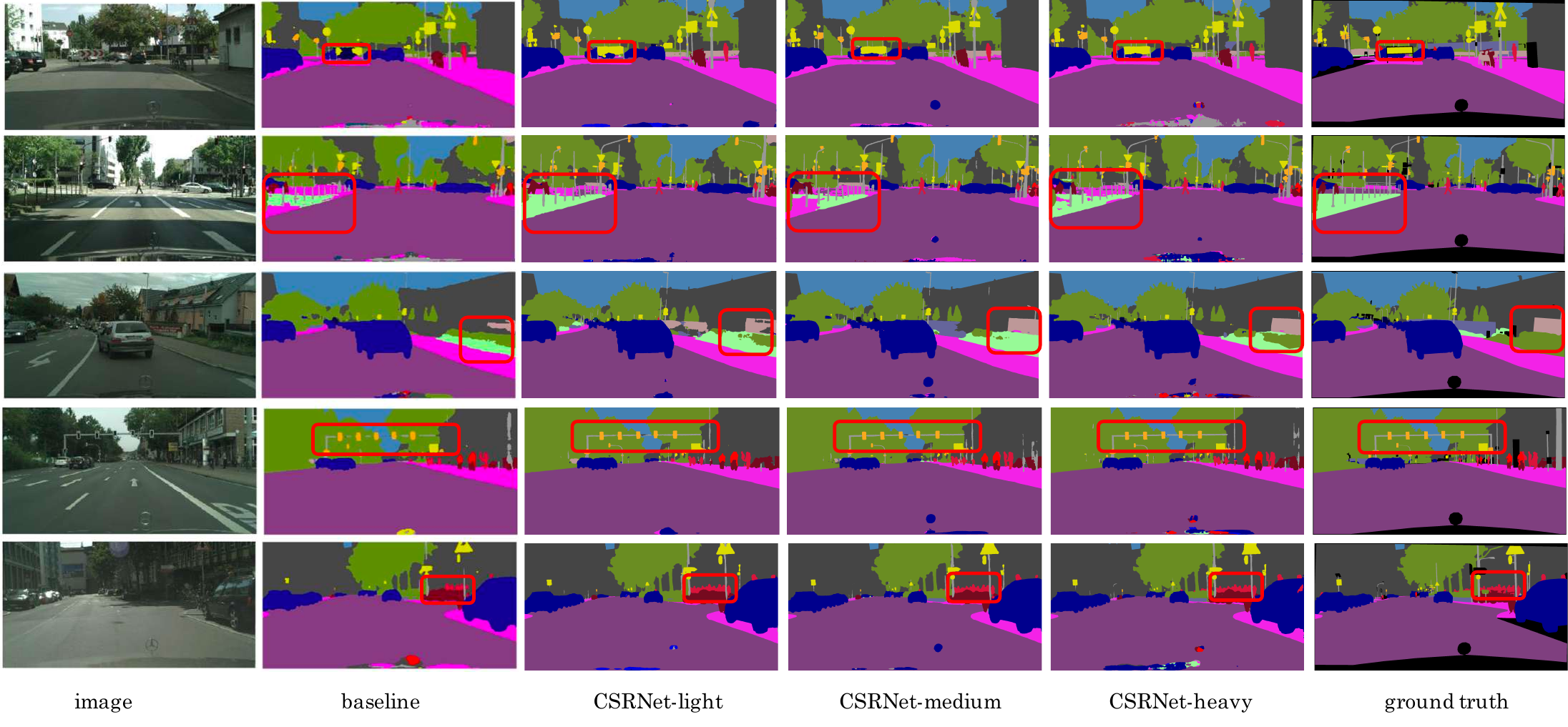}}
\caption[width=\linewidth]{Visualization examples on the Cityscapes validation set. Best viewed in color. Please zoom in for more details.}
\label{fig:example}
\end{figure}

\begin{figure}[t!]
\centering
{\includegraphics[width=\textwidth]{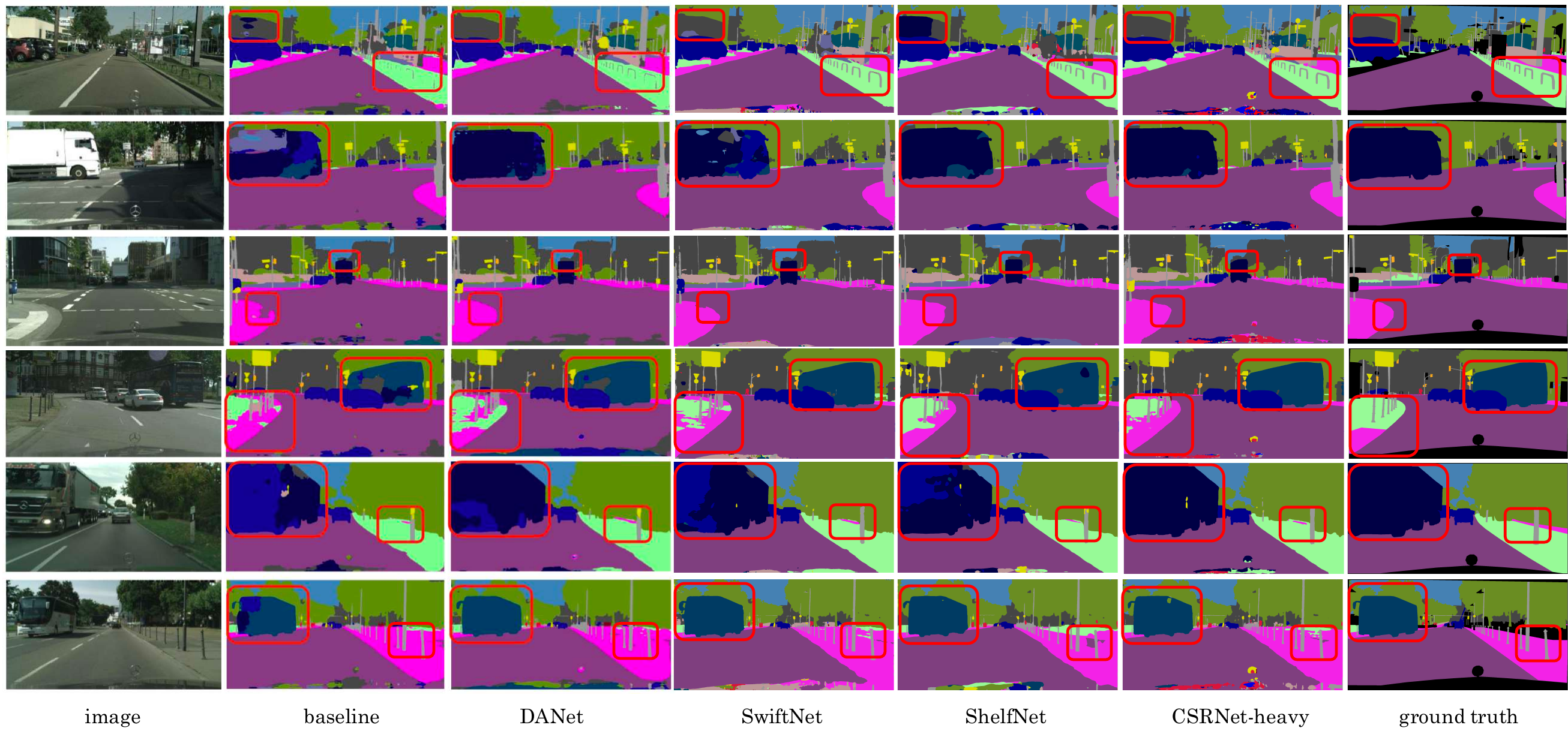}}
\caption[width=\linewidth]{Performances on segmenting thin, dominant or similar objects on the Cityscapes validation set. Best viewed in color. Please zoom in for more details.}
\label{fig:dominant}
\end{figure}

\subsubsection{Performances on Cityscapes Validation Set}
Some visual examples of CSRNet-heavy, CSRNet-medium, CSRNet-light on Cityscapes validation set are shown in Figure \ref{fig:example}. From this figure, we find that the segmentation qualities of the proposed networks are quite well, and the segmentation predictions become more and more convincing as the network becomes heavier. CSRNet-heavy has reached the most accurate results about the majority of objects regardless of their sizes, such as the “traffic sign” in the first row, the “truck” in the second row, the “fence” in the third row, the “pole” in the fourth row, and the “person” in the last row. 

\subsubsection{Performances on Segmenting Thin, Dominant or Similar Objects}
Comparison experiments of the segmentation performances on thin, dominant or similar objects with baseline, DANet \citep{fu2019dual}, SwiftNet \citep{orsic2019defense}, ShelfNet \citep{zhuang2019shelfnet} are conducted. DANet is a state-of-the-art semantic segmentation method that models the semantic interdependencies in spatial and channel dimensions. SwiftNet and ShelfNet are methods for real-time semantic segmentation. The visualization results are shown in Figure 3.13. The first to the third column are duplicated from DANet, which are the original input images from the Cityscapes validation set, segmentation results of baseline, and segmentation results of DANet, respectively. The fourth to the seventh column are the segmentation results of SwiftNet, ShelfNet, the proposed CSRNet-heavy and the ground truth.

As shown in Figure \ref{fig:dominant}, it is obvious that DANet, SwiftNet and ShelfNet achieve better performances on dominant objects than the baseline method. However, misclassification situations still exist in the interior of some dominant objects, such as the “bus” in the fourth row and the “truck” in the fifth row. Meanwhile, the segmentation of small objects can be further improved, like the “pole” in the first row. The prediction for pole pixels is discontinuous and thinner compared with the ground truth. Moreover, the previous methods also misclassify some truck pixels into car pixel. On the contrary, CSRNet-heavy has alleviated these problems with larger receptive fields and more comprehensive feature maps. It achieves high semantic consistency in dominant objects, and the object boundaries are more clear and precise than other methods. CSRNet has a cascaded network structure to progressively refine the feature map. In the feature aggregation process, CSRNet incorporates global semantics and preserves edge details.

\subsection{Ablation Study}
This subsection introduces the ablation experiments to prove the effectiveness of different components in the proposed network. The following experiments were conducted with ResNet-18 as the backbone and trained on the Cityscapes training set and evaluated on the Cityscapes validation set. 

\subsubsection{Ablation Study of Simplified Networks}
As discussed in Section \ref{sec:method}, we remove some modules in the heavy network and propose two simple networks to boost the inference time. The accuracy comparisons of mean IoU, running Time and FPS are reported in Table \ref{table:ablation} (\#\textbf{3}, \#\textbf{9}, \#\textbf{13}). mIoU (val) indicates the mean IoU of Cityscapes validation set. CSRNet-heavy achieves 77.33 mIoU, which outperforms the medium and light network version by 0.78\% and 1.22\%. Although the segmentation accuracies of light networks are slightly lower than the heavy network, they achieve a better balance between accuracy and speed. CSRNet-medium and CSRNet-light achieve a speed of 52.5 and 56 FPS, respectively, which are 1.55 - 1.65 times FPS of the heavy network.

\begin{table}[t]
	\renewcommand{\arraystretch}{0.9}
	\caption{Ablation Study of CSRNet-heavy, CSRNet-medium and CSRNet-light on Cityscapes dataset. Numbers in bold represent the best results.}
	\centering
	\label{table:ablation}
	\resizebox{\columnwidth}{!}{
		\begin{tabular}{cccccc}
        \hline
        Model architecture & Experiment number & Discription & mIoU (val) & Time (ms) & Frame (FPS)\\
        \hline
        \multirow{8}{*}{CSRNet-heavy} & \#\textbf{1} & Heavy-low-refined & 75.95 & 16.6 & 60.2\\
        & \#\textbf{2} & Heavy-mid-refined & 76.63 & 23.3 & 42.9\\
        & \#\textbf{3} & Heavy-high-refined (full model) & \textbf{77.33} & 27.5 & 36.3\\
        & \#\textbf{4} & Heavy-high-refined-without SPFM & 75.67 & 25.7 & 38.9\\
        & \#\textbf{5} & Heavy-low-refined-without attention & 75.71 & \textbf{15.3} & \textbf{65.4}\\
        & \#\textbf{6} & Heavy-mid-refined-without attention & 76.06 & 20.9 & 47.9\\
        & \#\textbf{7} & Heavy-high-refined-without attention & 76.17 & 24.1 & 41.5\\
        & \#\textbf{8} & Heavy-without 1 $\times$ 1 Conv in SRM & 76.32 & 24.9 & 40.1\\
        \hline
        \multirow{4}{*}{CSRNet-medium} & \#\textbf{9} & Medium (full model) & \textbf{76.55} & 19.0 & 52.5\\
        & \#\textbf{10} & Medium-without SPFM & 75.32 & \textbf{17.4} & \textbf{57.6}\\
        & \#\textbf{11} & Medium-without attention & 75.86 & 17.5 & 57.0\\
        & \#\textbf{12} & Medium-without 1 $\times$ 1 Conv in SRM & 76.05 & 18.0 & 55.7\\
        \hline
        \multirow{4}{*}{CSRNet-light} & \#\textbf{13} & Light (full model) & \textbf{76.11} & 17.9 & 56.0\\
        & \#\textbf{14} & Light-without SPFM & 74.89 & \textbf{16.2} & \textbf{61.8}\\
        & \#\textbf{15} & Light-without attention & 75.54 & 16.6 & 60.2\\
        & \#\textbf{16} & Light-without 1 $\times$ 1 Conv in SRM & 75.59 & 17.0 & 58.7\\
        \hline
		\end{tabular}
	}
\end{table}

\subsubsection{Ablation Study of Cascaded Stages}
The proposed heavy network builds three similar stages, where each stage starts the feature aggregation process from different path. The output features from Stage-1 to Stage-3 are called low-refined feature map, mid-refined feature map, and high-refined feature map, respectively, with the same spatial resolution and channel numbers. Based on the structure of CSRNet-heavy, we construct two sub-structure networks to quantitatively evaluate the feature refinement process. One design ends at Stage-1 and outputs the low-refined feature map as the final feature map, following a 1 $\times$ 1 convolution to make the final prediction (\#\textbf{1}). The other network ends at Stage-2 and outputs the mid-refined feature map as the final feature map, following a 1 $\times$ 1 convolution to make the prediction (\#\textbf{2}). \#\textbf{3} is the original heavy network. As can be seen from Experiment \#\textbf{1}, \#\textbf{2}, and \#\textbf{3} in Table \ref{table:ablation}, employing more stages result in higher segmentation accuracy (75.95 mIoU vs. 76.63 mIoU vs. 77.33 mIoU), but it will increase the inference time (60.2 FPS vs. 42.9 FPS vs. 36.3 FPS). 

\subsubsection{Ablation Study of SPFM}
The SPFM is responsible for incorporating the global context information and enlarging the receptive field. To clarify the effect of SPFM on the whole network, we replace SPFM with a 1 $\times$ 1 convolution followed by a batch-normalization layer and a ReLU layer. In Table \ref{table:ablation}, \#\textbf{4}, \#\textbf{10}, \#\textbf{14} are the counterparts of \#\textbf{3}, \#\textbf{9}, \#\textbf{13}, respectively. The SPFM improves the mean IoU in all experiments by approximately 1.3\% without sacrificing too much inference speed, proving that SPFM is an effective and efficient module in the proposed network.

\subsubsection{Ablation Study of SRM}
The SRM is designed to aggregate the feature maps with different resolutions and receptive fields. Compared with upsampling the lower feature map and summing the feature maps directly, the SRM exploits the attention mechanism, assigning different weights to the feature maps to adaptively select the feature receptive fields. In Table \ref{table:ablation}, \#\textbf{5}, \#\textbf{6}, \#\textbf{7}, \#\textbf{11}, \#\textbf{15} are the counterparts of \#\textbf{1}, \#\textbf{2}, \#\textbf{3}, \#\textbf{9}, \#\textbf{13} when we replace the attention mechanism in SRM with a simple addition operation (which is equivalent to $\mathbf{H}^{'}=\mathbf{H}$ and $\mathbf{L}^{'}=\mathcal{F}_{up}(\mathbf{L})$ in Figure \ref{fig:srm}). As shown in Table \ref{table:ablation}, SRM brings additional calculations, resulting in a slight FPS drop. However, it leads to segmentation accuracy gain in all experiments, especially for \#\textbf{3} vs. \#\textbf{7}, showing an increase of 0.89\% when we leverage SRM in CSRheavy-high-refined network. Moreover, adding the cascaded stages only brings a 0.46\% mIoU increase with the addition operation (\#\textbf{5} $\xrightarrow{}$ \#\textbf{7}), while the accuracy increment for SRM when applying cascaded stages is 1.11\% mIoU (\#\textbf{1} $\xrightarrow{}$ \#\textbf{3}). 

\begin{figure}[t]
\centering
{\includegraphics[width=\textwidth]{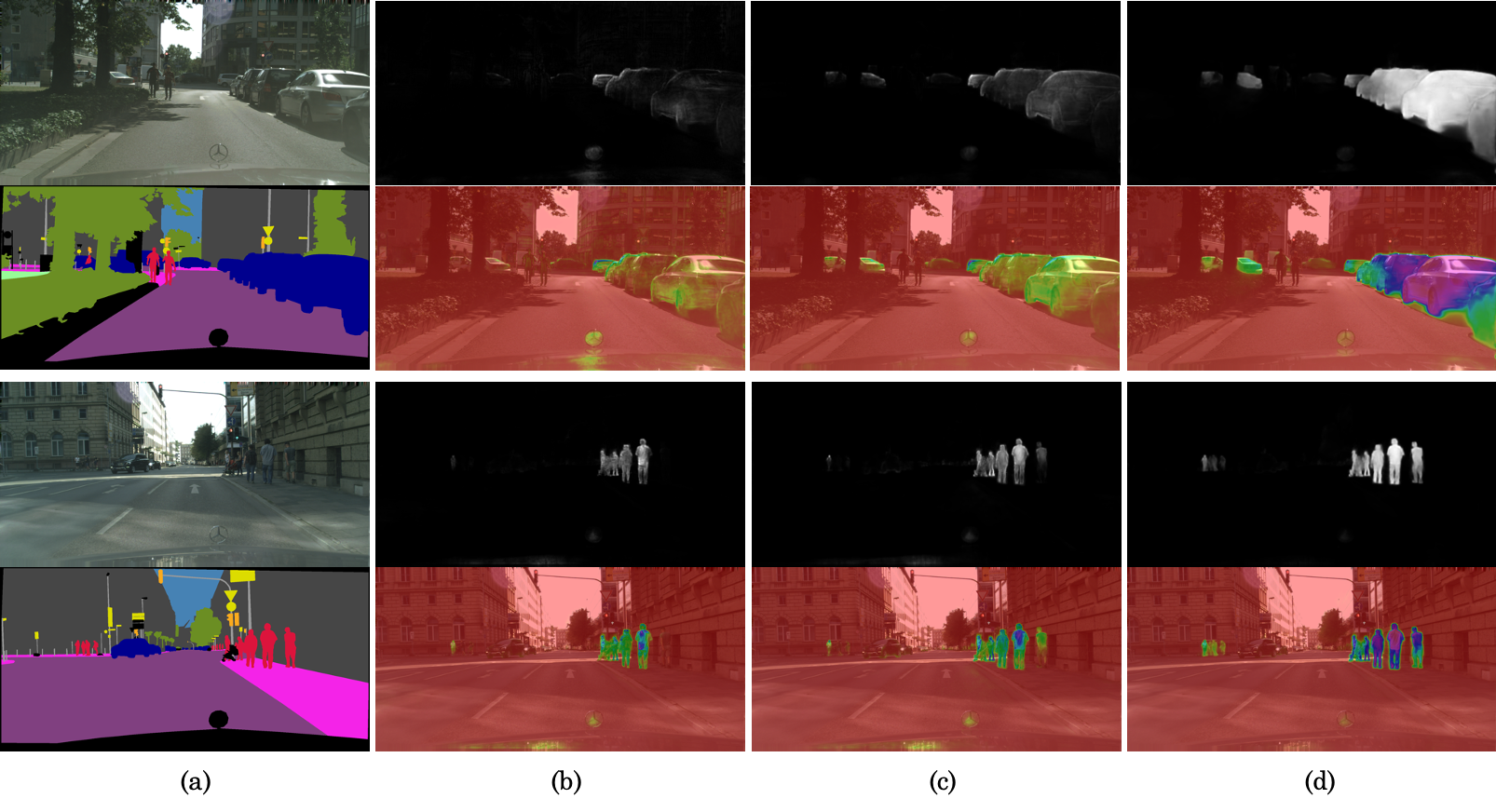}}
\caption[width=\linewidth]{Visualization of feature refinement process on Cityscapes dataset. (a) The input images and the ground truth. (b) The Score-CAM result of the low-refined feature map. (c) The Score-CAM result of the mid-refined feature map. (d) The Score-CAM result of the high-refined feature map. }
\label{fig:cam-citys}
\end{figure}

\begin{figure}[t]
\centering
{\includegraphics[width=\textwidth]{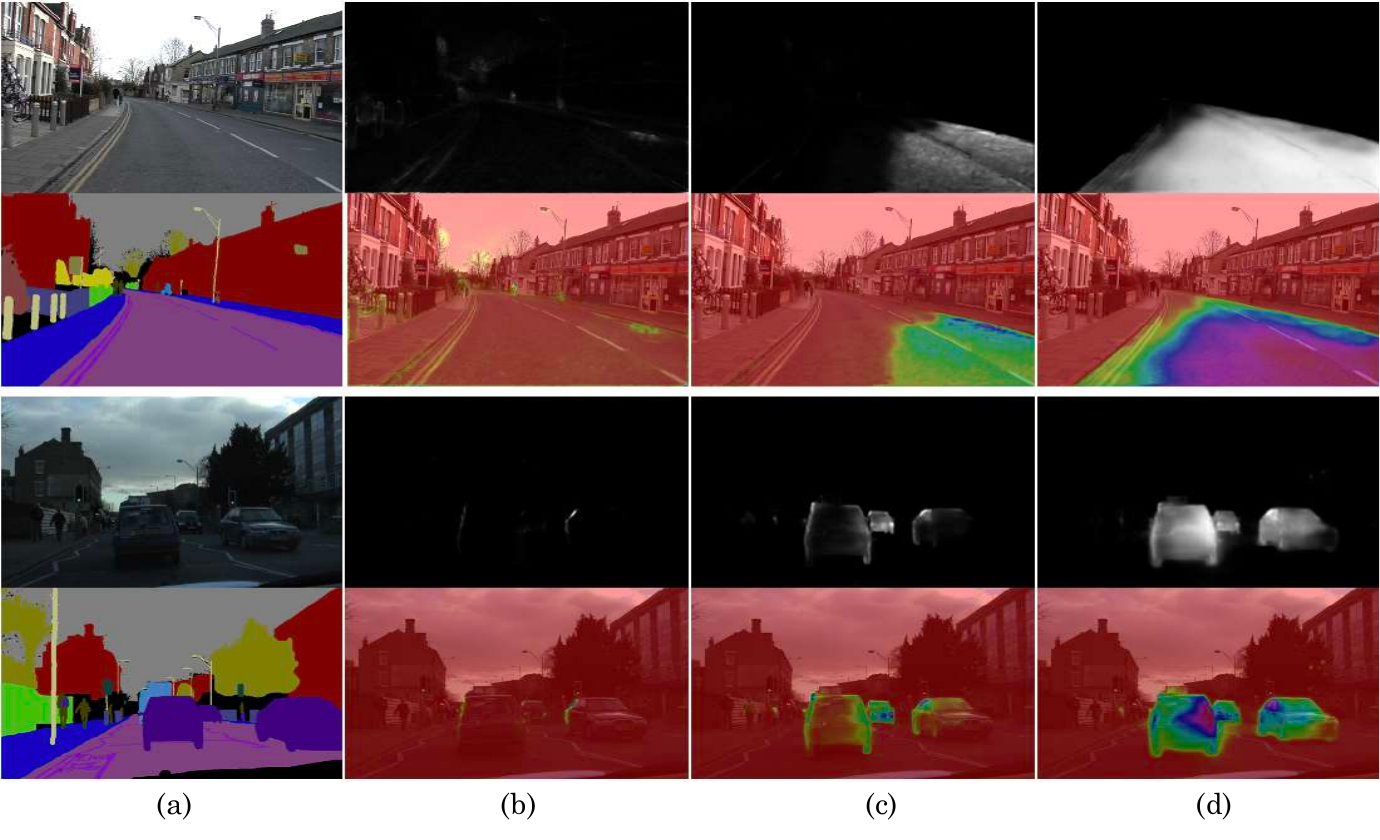}}
\caption[width=\linewidth]{Visualization of feature refinement process on CamVid dataset. (a) The input images and the ground truth. (b) The Score-CAM result of the low-refined feature map. (c) The Score-CAM result of the mid-refined feature map. (d) The Score-CAM result of the high-refined feature map. }
\label{fig:cam-camvid}
\end{figure}

\subsubsection{Ablation Study of the Convolutional Layer in SRM}
As discussed in Section \ref{sec:method}, at the end of SRM, it employs an 1 $\times$ 1 convolution and a 3 $\times$ 3 to blend the features. The 1 $\times$ 1 convolution is important to SRM since it aggregates channel information after merging the two feature maps into one. In Table \ref{table:ablation}, \#\textbf{8}, \#\textbf{12}, \#\textbf{16} are the counterparts of \#\textbf{3}, \#\textbf{9}, \#\textbf{13}, respectively where we delete the 1 $\times$ 1 convolution in SRM and keep other parts intact. From this table, we observe that removing the 1 $\times$ 1 convolution causes decreased accuracy in all experiments.

\subsection{Visualization of the Feature Refinement Process}
The class activation map (CAM) of a particular class represents the distinguished regions utilized by CNN to identify the class. This section exploits the Score-CAM method proposed in \citep{wang2020score} to generate the class activation map. The normalized activation map in Score-CAM works as a mask on the original input image, then a forward-passing score on the target class is obtained as the weight of the corresponding activation map. The Score-CAM is generated by a linear combination of score-based weights and activation maps. We employ Score-CAM at different convolutional layers in CSRNet-heavy for the target class “car” and “person” of two examples in the Cityscapes validation set. The visualization examples are shown in Figure \ref{fig:cam-citys}, demonstrating how localization regions change qualitatively when performing Score-CAM to the outputted feature maps in Stage-1 to Stage-3. We find that the high-refined feature map obtains the best-looking visualizations, and localizations get progressively worse at the early stages. Some small objects cannot be localized by the early-stage feature maps, e.g., the cars under the trees on the left side of the road in the first row, and people under the left building in the third row. Besides, the boundary in the last column is clearer than the first two columns. The high-refined   feature map captures more semantic concepts together with more spatial details. Similarly, the CAMs at different convolutional layers for the target class “road” and “car” of two examples in the CamVid testing set are shown in Figure \ref{fig:cam-camvid}. 

\section{Conclusion} \label{sec:conclu}
To fully investigate the feature fusion process of multi-resolution features with diverse receptive field, we propose a novel three-stage network CSRNet in this paper. The proposed network exploits the attention mechanism to aggregate feature maps with diverse spatial resolutions and receptive fields. Meanwhile, multiple adoption of global semantics embedding significantly enlarges the receptive field. Each stage aggregates feature information from the lower resolution to the higher resolution, and continues to incorporate context information and preserve edge details. The multi-stage process can be regarded as a feature refinement process, and the information flow between different stages is flexible. The output feature maps of each stage have the same spatial size and dimension, while the feature maps become more and more discriminative and recognizable. Extensive experiments demonstrate that the proposed CSRNet effectively improves the performance for real-time segmentation. We hope this work will foster further research in real-time semantic segmentation. In future work, we will extend the proposed network to efficient video semantic segmentation and exploit the spatial consistency clues in videos to ensure the speed and accuracy of segmentation.

\bibliography{csrnet}

\begin{thebibliography}{61}
\expandafter\ifx\csname natexlab\endcsname\relax\def\natexlab#1{#1}\fi
\providecommand{\url}[1]{\texttt{#1}}
\providecommand{\href}[2]{#2}
\providecommand{\path}[1]{#1}
\providecommand{\DOIprefix}{doi:}
\providecommand{\ArXivprefix}{arXiv:}
\providecommand{\URLprefix}{URL: }
\providecommand{\Pubmedprefix}{pmid:}
\providecommand{\doi}[1]{\href{http://dx.doi.org/#1}{\path{#1}}}
\providecommand{\Pubmed}[1]{\href{pmid:#1}{\path{#1}}}
\providecommand{\bibinfo}[2]{#2}
\ifx\xfnm\relax \def\xfnm[#1]{\unskip,\space#1}\fi
\bibitem[{Arani et~al.(2021)Arani, Marzban, Pata \& Zonooz}]{arani2021rgpnet}
\bibinfo{author}{Arani, E.}, \bibinfo{author}{Marzban, S.},
  \bibinfo{author}{Pata, A.}, \& \bibinfo{author}{Zonooz, B.}
  (\bibinfo{year}{2021}).
\newblock \bibinfo{title}{Rgpnet: A real-time general purpose semantic
  segmentation}.
\newblock In {\it \bibinfo{booktitle}{Proceedings of the IEEE/CVF Winter
  Conference on Applications of Computer Vision}\/} (pp.
  \bibinfo{pages}{3009--3018}).
\bibitem[{Badrinarayanan et~al.(2017)Badrinarayanan, Kendall \&
  Cipolla}]{badrinarayanan2017segnet}
\bibinfo{author}{Badrinarayanan, V.}, \bibinfo{author}{Kendall, A.}, \&
  \bibinfo{author}{Cipolla, R.} (\bibinfo{year}{2017}).
\newblock \bibinfo{title}{Segnet: A deep convolutional encoder-decoder
  architecture for image segmentation}.
\newblock {\it \bibinfo{journal}{IEEE transactions on pattern analysis and
  machine intelligence}\/},  {\it \bibinfo{volume}{39}\/},
  \bibinfo{pages}{2481--2495}.
\bibitem[{Bahdanau et~al.(2014)Bahdanau, Cho \& Bengio}]{bahdanau2014neural}
\bibinfo{author}{Bahdanau, D.}, \bibinfo{author}{Cho, K.}, \&
  \bibinfo{author}{Bengio, Y.} (\bibinfo{year}{2014}).
\newblock \bibinfo{title}{Neural machine translation by jointly learning to
  align and translate}.
\newblock {\it \bibinfo{journal}{arXiv preprint arXiv:1409.0473}\/}, .
\bibitem[{Brostow et~al.(2008)Brostow, Shotton, Fauqueur \&
  Cipolla}]{brostow2008segmentation}
\bibinfo{author}{Brostow, G.~J.}, \bibinfo{author}{Shotton, J.},
  \bibinfo{author}{Fauqueur, J.}, \& \bibinfo{author}{Cipolla, R.}
  (\bibinfo{year}{2008}).
\newblock \bibinfo{title}{Segmentation and recognition using structure from
  motion point clouds}.
\newblock In {\it \bibinfo{booktitle}{European conference on computer
  vision}\/} (pp. \bibinfo{pages}{44--57}).
\newblock \bibinfo{organization}{Springer}.
\bibitem[{Cao et~al.(2019)Cao, Xu, Lin, Wei \& Hu}]{cao2019gcnet}
\bibinfo{author}{Cao, Y.}, \bibinfo{author}{Xu, J.}, \bibinfo{author}{Lin, S.},
  \bibinfo{author}{Wei, F.}, \& \bibinfo{author}{Hu, H.}
  (\bibinfo{year}{2019}).
\newblock \bibinfo{title}{Gcnet: Non-local networks meet squeeze-excitation
  networks and beyond}.
\newblock In {\it \bibinfo{booktitle}{Proceedings of the IEEE/CVF International
  Conference on Computer Vision Workshops}\/} (pp. \bibinfo{pages}{0--0}).
\bibitem[{Chen et~al.(2017{\natexlab{a}})Chen, Papandreou, Kokkinos, Murphy \&
  Yuille}]{chen2017deeplab}
\bibinfo{author}{Chen, L.-C.}, \bibinfo{author}{Papandreou, G.},
  \bibinfo{author}{Kokkinos, I.}, \bibinfo{author}{Murphy, K.}, \&
  \bibinfo{author}{Yuille, A.~L.} (\bibinfo{year}{2017}{\natexlab{a}}).
\newblock \bibinfo{title}{Deeplab: Semantic image segmentation with deep
  convolutional nets, atrous convolution, and fully connected crfs}.
\newblock {\it \bibinfo{journal}{IEEE transactions on pattern analysis and
  machine intelligence}\/},  {\it \bibinfo{volume}{40}\/},
  \bibinfo{pages}{834--848}.
\bibitem[{Chen et~al.(2017{\natexlab{b}})Chen, Papandreou, Schroff \&
  Adam}]{chen2017rethinking}
\bibinfo{author}{Chen, L.-C.}, \bibinfo{author}{Papandreou, G.},
  \bibinfo{author}{Schroff, F.}, \& \bibinfo{author}{Adam, H.}
  (\bibinfo{year}{2017}{\natexlab{b}}).
\newblock \bibinfo{title}{Rethinking atrous convolution for semantic image
  segmentation}.
\newblock {\it \bibinfo{journal}{arXiv preprint arXiv:1706.05587}\/}, .
\bibitem[{Chen et~al.(2018)Chen, Kalantidis, Li, Yan \& Feng}]{chen20182}
\bibinfo{author}{Chen, Y.}, \bibinfo{author}{Kalantidis, Y.},
  \bibinfo{author}{Li, J.}, \bibinfo{author}{Yan, S.}, \&
  \bibinfo{author}{Feng, J.} (\bibinfo{year}{2018}).
\newblock \bibinfo{title}{A2-nets: Double attention networks}.
\newblock {\it \bibinfo{journal}{arXiv preprint arXiv:1810.11579}\/}, .
\bibitem[{Chollet(2017)}]{chollet2017xception}
\bibinfo{author}{Chollet, F.} (\bibinfo{year}{2017}).
\newblock \bibinfo{title}{Xception: Deep learning with depthwise separable
  convolutions}.
\newblock In {\it \bibinfo{booktitle}{Proceedings of the IEEE conference on
  computer vision and pattern recognition}\/} (pp.
  \bibinfo{pages}{1251--1258}).
\bibitem[{Cordts et~al.(2016)Cordts, Omran, Ramos, Rehfeld, Enzweiler,
  Benenson, Franke, Roth \& Schiele}]{cordts2016cityscapes}
\bibinfo{author}{Cordts, M.}, \bibinfo{author}{Omran, M.},
  \bibinfo{author}{Ramos, S.}, \bibinfo{author}{Rehfeld, T.},
  \bibinfo{author}{Enzweiler, M.}, \bibinfo{author}{Benenson, R.},
  \bibinfo{author}{Franke, U.}, \bibinfo{author}{Roth, S.}, \&
  \bibinfo{author}{Schiele, B.} (\bibinfo{year}{2016}).
\newblock \bibinfo{title}{The cityscapes dataset for semantic urban scene
  understanding}.
\newblock In {\it \bibinfo{booktitle}{Proceedings of the IEEE conference on
  computer vision and pattern recognition}\/} (pp.
  \bibinfo{pages}{3213--3223}).
\bibitem[{Dong et~al.(2020)Dong, Yan, Shen \& Wang}]{dong2020real}
\bibinfo{author}{Dong, G.}, \bibinfo{author}{Yan, Y.}, \bibinfo{author}{Shen,
  C.}, \& \bibinfo{author}{Wang, H.} (\bibinfo{year}{2020}).
\newblock \bibinfo{title}{Real-time high-performance semantic image
  segmentation of urban street scenes}.
\newblock {\it \bibinfo{journal}{IEEE Transactions on Intelligent
  Transportation Systems}\/}, .
\bibitem[{Elhassan et~al.(2021)Elhassan, Huang, Yang \&
  Munea}]{elhassan2021dsanet}
\bibinfo{author}{Elhassan, M.~A.}, \bibinfo{author}{Huang, C.},
  \bibinfo{author}{Yang, C.}, \& \bibinfo{author}{Munea, T.~L.}
  (\bibinfo{year}{2021}).
\newblock \bibinfo{title}{Dsanet: Dilated spatial attention for real-time
  semantic segmentation in urban street scenes}.
\newblock {\it \bibinfo{journal}{Expert Systems with Applications}\/},  (p.
  \bibinfo{pages}{115090}).
\bibitem[{Fu et~al.(2020)Fu, Liu, Jiang, Li, Bao \& Lu}]{fu2020scene}
\bibinfo{author}{Fu, J.}, \bibinfo{author}{Liu, J.}, \bibinfo{author}{Jiang,
  J.}, \bibinfo{author}{Li, Y.}, \bibinfo{author}{Bao, Y.}, \&
  \bibinfo{author}{Lu, H.} (\bibinfo{year}{2020}).
\newblock \bibinfo{title}{Scene segmentation with dual relation-aware attention
  network}.
\newblock {\it \bibinfo{journal}{IEEE Transactions on Neural Networks and
  Learning Systems}\/}, .
\bibitem[{Fu et~al.(2019)Fu, Liu, Tian, Li, Bao, Fang \& Lu}]{fu2019dual}
\bibinfo{author}{Fu, J.}, \bibinfo{author}{Liu, J.}, \bibinfo{author}{Tian,
  H.}, \bibinfo{author}{Li, Y.}, \bibinfo{author}{Bao, Y.},
  \bibinfo{author}{Fang, Z.}, \& \bibinfo{author}{Lu, H.}
  (\bibinfo{year}{2019}).
\newblock \bibinfo{title}{Dual attention network for scene segmentation}.
\newblock In {\it \bibinfo{booktitle}{Proceedings of the IEEE/CVF Conference on
  Computer Vision and Pattern Recognition}\/} (pp.
  \bibinfo{pages}{3146--3154}).
\bibitem[{He et~al.(2019)He, Deng, Zhou, Wang \& Qiao}]{he2019adaptive}
\bibinfo{author}{He, J.}, \bibinfo{author}{Deng, Z.}, \bibinfo{author}{Zhou,
  L.}, \bibinfo{author}{Wang, Y.}, \& \bibinfo{author}{Qiao, Y.}
  (\bibinfo{year}{2019}).
\newblock \bibinfo{title}{Adaptive pyramid context network for semantic
  segmentation}.
\newblock In {\it \bibinfo{booktitle}{Proceedings of the IEEE/CVF Conference on
  Computer Vision and Pattern Recognition}\/} (pp.
  \bibinfo{pages}{7519--7528}).
\bibitem[{He et~al.(2016)He, Zhang, Ren \& Sun}]{he2016deep}
\bibinfo{author}{He, K.}, \bibinfo{author}{Zhang, X.}, \bibinfo{author}{Ren,
  S.}, \& \bibinfo{author}{Sun, J.} (\bibinfo{year}{2016}).
\newblock \bibinfo{title}{Deep residual learning for image recognition}.
\newblock In {\it \bibinfo{booktitle}{Proceedings of the IEEE conference on
  computer vision and pattern recognition}\/} (pp. \bibinfo{pages}{770--778}).
\bibitem[{Howard et~al.(2019)Howard, Sandler, Chu, Chen, Chen, Tan, Wang, Zhu,
  Pang, Vasudevan et~al.}]{howard2019searching}
\bibinfo{author}{Howard, A.}, \bibinfo{author}{Sandler, M.},
  \bibinfo{author}{Chu, G.}, \bibinfo{author}{Chen, L.-C.},
  \bibinfo{author}{Chen, B.}, \bibinfo{author}{Tan, M.}, \bibinfo{author}{Wang,
  W.}, \bibinfo{author}{Zhu, Y.}, \bibinfo{author}{Pang, R.},
  \bibinfo{author}{Vasudevan, V.} et~al. (\bibinfo{year}{2019}).
\newblock \bibinfo{title}{Searching for mobilenetv3}.
\newblock In {\it \bibinfo{booktitle}{Proceedings of the IEEE/CVF International
  Conference on Computer Vision}\/} (pp. \bibinfo{pages}{1314--1324}).
\bibitem[{Howard et~al.(2017)Howard, Zhu, Chen, Kalenichenko, Wang, Weyand,
  Andreetto \& Adam}]{howard2017mobilenets}
\bibinfo{author}{Howard, A.~G.}, \bibinfo{author}{Zhu, M.},
  \bibinfo{author}{Chen, B.}, \bibinfo{author}{Kalenichenko, D.},
  \bibinfo{author}{Wang, W.}, \bibinfo{author}{Weyand, T.},
  \bibinfo{author}{Andreetto, M.}, \& \bibinfo{author}{Adam, H.}
  (\bibinfo{year}{2017}).
\newblock \bibinfo{title}{Mobilenets: Efficient convolutional neural networks
  for mobile vision applications}.
\newblock {\it \bibinfo{journal}{arXiv preprint arXiv:1704.04861}\/}, .
\bibitem[{Hu et~al.(2018)Hu, Shen \& Sun}]{hu2018squeeze}
\bibinfo{author}{Hu, J.}, \bibinfo{author}{Shen, L.}, \& \bibinfo{author}{Sun,
  G.} (\bibinfo{year}{2018}).
\newblock \bibinfo{title}{Squeeze-and-excitation networks}.
\newblock In {\it \bibinfo{booktitle}{Proceedings of the IEEE conference on
  computer vision and pattern recognition}\/} (pp.
  \bibinfo{pages}{7132--7141}).
\bibitem[{Hu et~al.(2020)Hu, Perazzi, Heilbron, Wang, Lin, Saenko \&
  Sclaroff}]{hu2020real}
\bibinfo{author}{Hu, P.}, \bibinfo{author}{Perazzi, F.},
  \bibinfo{author}{Heilbron, F.~C.}, \bibinfo{author}{Wang, O.},
  \bibinfo{author}{Lin, Z.}, \bibinfo{author}{Saenko, K.}, \&
  \bibinfo{author}{Sclaroff, S.} (\bibinfo{year}{2020}).
\newblock \bibinfo{title}{Real-time semantic segmentation with fast attention}.
\newblock {\it \bibinfo{journal}{IEEE Robotics and Automation Letters}\/},
  {\it \bibinfo{volume}{6}\/}, \bibinfo{pages}{263--270}.
\bibitem[{Huang et~al.(2019{\natexlab{a}})Huang, Yuan, Guo, Zhang, Chen \&
  Wang}]{huang2019interlaced}
\bibinfo{author}{Huang, L.}, \bibinfo{author}{Yuan, Y.}, \bibinfo{author}{Guo,
  J.}, \bibinfo{author}{Zhang, C.}, \bibinfo{author}{Chen, X.}, \&
  \bibinfo{author}{Wang, J.} (\bibinfo{year}{2019}{\natexlab{a}}).
\newblock \bibinfo{title}{Interlaced sparse self-attention for semantic
  segmentation}.
\newblock {\it \bibinfo{journal}{arXiv preprint arXiv:1907.12273}\/}, .
\bibitem[{Huang et~al.(2019{\natexlab{b}})Huang, Wang, Huang, Huang, Wei \&
  Liu}]{huang2019ccnet}
\bibinfo{author}{Huang, Z.}, \bibinfo{author}{Wang, X.},
  \bibinfo{author}{Huang, L.}, \bibinfo{author}{Huang, C.},
  \bibinfo{author}{Wei, Y.}, \& \bibinfo{author}{Liu, W.}
  (\bibinfo{year}{2019}{\natexlab{b}}).
\newblock \bibinfo{title}{Ccnet: Criss-cross attention for semantic
  segmentation}.
\newblock In {\it \bibinfo{booktitle}{Proceedings of the IEEE/CVF International
  Conference on Computer Vision}\/} (pp. \bibinfo{pages}{603--612}).
\bibitem[{Ilya \& Hutter(2016)}]{ilya2016sgdr}
\bibinfo{author}{Ilya, L.}, \& \bibinfo{author}{Hutter, F.}
  (\bibinfo{year}{2016}).
\newblock \bibinfo{title}{Sgdr: stochastic gradient descent with restarts}.
\newblock {\it \bibinfo{journal}{Learning}\/},  {\it \bibinfo{volume}{10}\/}.
\bibitem[{Ioffe \& Szegedy(2015)}]{ioffe2015batch}
\bibinfo{author}{Ioffe, S.}, \& \bibinfo{author}{Szegedy, C.}
  (\bibinfo{year}{2015}).
\newblock \bibinfo{title}{Batch normalization: Accelerating deep network
  training by reducing internal covariate shift}.
\newblock In {\it \bibinfo{booktitle}{International conference on machine
  learning}\/} (pp. \bibinfo{pages}{448--456}).
\newblock \bibinfo{organization}{PMLR}.
\bibitem[{Kingma \& Ba(2014)}]{kingma2014adam}
\bibinfo{author}{Kingma, D.~P.}, \& \bibinfo{author}{Ba, J.}
  (\bibinfo{year}{2014}).
\newblock \bibinfo{title}{Adam: A method for stochastic optimization}.
\newblock {\it \bibinfo{journal}{arXiv preprint arXiv:1412.6980}\/}, .
\bibitem[{Li et~al.(2019{\natexlab{a}})Li, Yun, Kim \& Kim}]{li2019dabnet}
\bibinfo{author}{Li, G.}, \bibinfo{author}{Yun, I.}, \bibinfo{author}{Kim, J.},
  \& \bibinfo{author}{Kim, J.} (\bibinfo{year}{2019}{\natexlab{a}}).
\newblock \bibinfo{title}{Dabnet: Depth-wise asymmetric bottleneck for
  real-time semantic segmentation}.
\newblock {\it \bibinfo{journal}{arXiv preprint arXiv:1907.11357}\/}, .
\bibitem[{Li et~al.(2019{\natexlab{b}})Li, Xiong, Fan \& Sun}]{li2019dfanet}
\bibinfo{author}{Li, H.}, \bibinfo{author}{Xiong, P.}, \bibinfo{author}{Fan,
  H.}, \& \bibinfo{author}{Sun, J.} (\bibinfo{year}{2019}{\natexlab{b}}).
\newblock \bibinfo{title}{Dfanet: Deep feature aggregation for real-time
  semantic segmentation}.
\newblock In {\it \bibinfo{booktitle}{Proceedings of the IEEE/CVF Conference on
  Computer Vision and Pattern Recognition}\/} (pp.
  \bibinfo{pages}{9522--9531}).
\bibitem[{Li et~al.(2019{\natexlab{c}})Li, Wang, Hu \& Yang}]{li2019selective}
\bibinfo{author}{Li, X.}, \bibinfo{author}{Wang, W.}, \bibinfo{author}{Hu, X.},
  \& \bibinfo{author}{Yang, J.} (\bibinfo{year}{2019}{\natexlab{c}}).
\newblock \bibinfo{title}{Selective kernel networks}.
\newblock In {\it \bibinfo{booktitle}{Proceedings of the IEEE/CVF Conference on
  Computer Vision and Pattern Recognition}\/} (pp. \bibinfo{pages}{510--519}).
\bibitem[{Li et~al.(2020)Li, You, Zhu, Zhao, Yang, Yang, Tan \&
  Tong}]{li2020semantic}
\bibinfo{author}{Li, X.}, \bibinfo{author}{You, A.}, \bibinfo{author}{Zhu, Z.},
  \bibinfo{author}{Zhao, H.}, \bibinfo{author}{Yang, M.},
  \bibinfo{author}{Yang, K.}, \bibinfo{author}{Tan, S.}, \&
  \bibinfo{author}{Tong, Y.} (\bibinfo{year}{2020}).
\newblock \bibinfo{title}{Semantic flow for fast and accurate scene parsing}.
\newblock In {\it \bibinfo{booktitle}{European Conference on Computer
  Vision}\/} (pp. \bibinfo{pages}{775--793}).
\newblock \bibinfo{organization}{Springer}.
\bibitem[{Li et~al.(2019{\natexlab{d}})Li, Zhong, Wu, Yang, Lin \&
  Liu}]{li2019expectation}
\bibinfo{author}{Li, X.}, \bibinfo{author}{Zhong, Z.}, \bibinfo{author}{Wu,
  J.}, \bibinfo{author}{Yang, Y.}, \bibinfo{author}{Lin, Z.}, \&
  \bibinfo{author}{Liu, H.} (\bibinfo{year}{2019}{\natexlab{d}}).
\newblock \bibinfo{title}{Expectation-maximization attention networks for
  semantic segmentation}.
\newblock In {\it \bibinfo{booktitle}{Proceedings of the IEEE/CVF International
  Conference on Computer Vision}\/} (pp. \bibinfo{pages}{9167--9176}).
\bibitem[{Li et~al.(2021)Li, Liu \& Sun}]{li2021real}
\bibinfo{author}{Li, Y.}, \bibinfo{author}{Liu, Y.}, \& \bibinfo{author}{Sun,
  Q.} (\bibinfo{year}{2021}).
\newblock \bibinfo{title}{Real-time semantic segmentation via region and pixel
  context network}.
\newblock In {\it \bibinfo{booktitle}{2020 25th International Conference on
  Pattern Recognition (ICPR)}\/} (pp. \bibinfo{pages}{7043--7049}).
\newblock \bibinfo{organization}{IEEE}.
\bibitem[{Lin et~al.(2020)Lin, Sun, Cheng, Xie, Li \& Shi}]{lin2020graph}
\bibinfo{author}{Lin, P.}, \bibinfo{author}{Sun, P.}, \bibinfo{author}{Cheng,
  G.}, \bibinfo{author}{Xie, S.}, \bibinfo{author}{Li, X.}, \&
  \bibinfo{author}{Shi, J.} (\bibinfo{year}{2020}).
\newblock \bibinfo{title}{Graph-guided architecture search for real-time
  semantic segmentation}.
\newblock In {\it \bibinfo{booktitle}{Proceedings of the IEEE/CVF Conference on
  Computer Vision and Pattern Recognition}\/} (pp.
  \bibinfo{pages}{4203--4212}).
\bibitem[{Luong et~al.(2015)Luong, Pham \& Manning}]{luong2015effective}
\bibinfo{author}{Luong, M.-T.}, \bibinfo{author}{Pham, H.}, \&
  \bibinfo{author}{Manning, C.~D.} (\bibinfo{year}{2015}).
\newblock \bibinfo{title}{Effective approaches to attention-based neural
  machine translation}.
\newblock {\it \bibinfo{journal}{arXiv preprint arXiv:1508.04025}\/}, .
\bibitem[{Mazzini(2018)}]{mazzini2018guided}
\bibinfo{author}{Mazzini, D.} (\bibinfo{year}{2018}).
\newblock \bibinfo{title}{Guided upsampling network for real-time semantic
  segmentation}.
\newblock {\it \bibinfo{journal}{arXiv preprint arXiv:1807.07466}\/}, .
\bibitem[{Mehta et~al.(2018)Mehta, Rastegari, Caspi, Shapiro \&
  Hajishirzi}]{mehta2018espnet}
\bibinfo{author}{Mehta, S.}, \bibinfo{author}{Rastegari, M.},
  \bibinfo{author}{Caspi, A.}, \bibinfo{author}{Shapiro, L.}, \&
  \bibinfo{author}{Hajishirzi, H.} (\bibinfo{year}{2018}).
\newblock \bibinfo{title}{Espnet: Efficient spatial pyramid of dilated
  convolutions for semantic segmentation}.
\newblock In {\it \bibinfo{booktitle}{Proceedings of the european conference on
  computer vision (ECCV)}\/} (pp. \bibinfo{pages}{552--568}).
\bibitem[{Mehta et~al.(2019)Mehta, Rastegari, Shapiro \&
  Hajishirzi}]{mehta2019espnetv2}
\bibinfo{author}{Mehta, S.}, \bibinfo{author}{Rastegari, M.},
  \bibinfo{author}{Shapiro, L.}, \& \bibinfo{author}{Hajishirzi, H.}
  (\bibinfo{year}{2019}).
\newblock \bibinfo{title}{Espnetv2: A light-weight, power efficient, and
  general purpose convolutional neural network}.
\newblock In {\it \bibinfo{booktitle}{Proceedings of the IEEE/CVF Conference on
  Computer Vision and Pattern Recognition}\/} (pp.
  \bibinfo{pages}{9190--9200}).
\bibitem[{Orsic et~al.(2019)Orsic, Kreso, Bevandic \&
  Segvic}]{orsic2019defense}
\bibinfo{author}{Orsic, M.}, \bibinfo{author}{Kreso, I.},
  \bibinfo{author}{Bevandic, P.}, \& \bibinfo{author}{Segvic, S.}
  (\bibinfo{year}{2019}).
\newblock \bibinfo{title}{In defense of pre-trained imagenet architectures for
  real-time semantic segmentation of road-driving images}.
\newblock In {\it \bibinfo{booktitle}{Proceedings of the IEEE/CVF Conference on
  Computer Vision and Pattern Recognition}\/} (pp.
  \bibinfo{pages}{12607--12616}).
\bibitem[{Or{\v{s}}i{\'c} \& {\v{S}}egvi{\'c}(2021)}]{orvsic2021efficient}
\bibinfo{author}{Or{\v{s}}i{\'c}, M.}, \& \bibinfo{author}{{\v{S}}egvi{\'c},
  S.} (\bibinfo{year}{2021}).
\newblock \bibinfo{title}{Efficient semantic segmentation with pyramidal
  fusion}.
\newblock {\it \bibinfo{journal}{Pattern Recognition}\/},  {\it
  \bibinfo{volume}{110}\/}, \bibinfo{pages}{107611}.
\bibitem[{Paszke et~al.(2016)Paszke, Chaurasia, Kim \&
  Culurciello}]{paszke2016enet}
\bibinfo{author}{Paszke, A.}, \bibinfo{author}{Chaurasia, A.},
  \bibinfo{author}{Kim, S.}, \& \bibinfo{author}{Culurciello, E.}
  (\bibinfo{year}{2016}).
\newblock \bibinfo{title}{Enet: A deep neural network architecture for
  real-time semantic segmentation}.
\newblock {\it \bibinfo{journal}{arXiv preprint arXiv:1606.02147}\/}, .
\bibitem[{Poudel et~al.(2019)Poudel, Liwicki \& Cipolla}]{poudel2019fast}
\bibinfo{author}{Poudel, R.~P.}, \bibinfo{author}{Liwicki, S.}, \&
  \bibinfo{author}{Cipolla, R.} (\bibinfo{year}{2019}).
\newblock \bibinfo{title}{Fast-scnn: fast semantic segmentation network}.
\newblock {\it \bibinfo{journal}{arXiv preprint arXiv:1902.04502}\/}, .
\bibitem[{Romera et~al.(2017)Romera, Alvarez, Bergasa \&
  Arroyo}]{romera2017erfnet}
\bibinfo{author}{Romera, E.}, \bibinfo{author}{Alvarez, J.~M.},
  \bibinfo{author}{Bergasa, L.~M.}, \& \bibinfo{author}{Arroyo, R.}
  (\bibinfo{year}{2017}).
\newblock \bibinfo{title}{Erfnet: Efficient residual factorized convnet for
  real-time semantic segmentation}.
\newblock {\it \bibinfo{journal}{IEEE Transactions on Intelligent
  Transportation Systems}\/},  {\it \bibinfo{volume}{19}\/},
  \bibinfo{pages}{263--272}.
\bibitem[{Sandler et~al.(2018)Sandler, Howard, Zhu, Zhmoginov \&
  Chen}]{sandler2018mobilenetv2}
\bibinfo{author}{Sandler, M.}, \bibinfo{author}{Howard, A.},
  \bibinfo{author}{Zhu, M.}, \bibinfo{author}{Zhmoginov, A.}, \&
  \bibinfo{author}{Chen, L.-C.} (\bibinfo{year}{2018}).
\newblock \bibinfo{title}{Mobilenetv2: Inverted residuals and linear
  bottlenecks}.
\newblock In {\it \bibinfo{booktitle}{Proceedings of the IEEE conference on
  computer vision and pattern recognition}\/} (pp.
  \bibinfo{pages}{4510--4520}).
\bibitem[{Szegedy et~al.(2015)Szegedy, Liu, Jia, Sermanet, Reed, Anguelov,
  Erhan, Vanhoucke \& Rabinovich}]{szegedy2015going}
\bibinfo{author}{Szegedy, C.}, \bibinfo{author}{Liu, W.}, \bibinfo{author}{Jia,
  Y.}, \bibinfo{author}{Sermanet, P.}, \bibinfo{author}{Reed, S.},
  \bibinfo{author}{Anguelov, D.}, \bibinfo{author}{Erhan, D.},
  \bibinfo{author}{Vanhoucke, V.}, \& \bibinfo{author}{Rabinovich, A.}
  (\bibinfo{year}{2015}).
\newblock \bibinfo{title}{Going deeper with convolutions}.
\newblock In {\it \bibinfo{booktitle}{Proceedings of the IEEE conference on
  computer vision and pattern recognition}\/} (pp. \bibinfo{pages}{1--9}).
\bibitem[{Tao et~al.(2020)Tao, Sapra \& Catanzaro}]{tao2020hierarchical}
\bibinfo{author}{Tao, A.}, \bibinfo{author}{Sapra, K.}, \&
  \bibinfo{author}{Catanzaro, B.} (\bibinfo{year}{2020}).
\newblock \bibinfo{title}{Hierarchical multi-scale attention for semantic
  segmentation}.
\newblock {\it \bibinfo{journal}{arXiv preprint arXiv:2005.10821}\/}, .
\bibitem[{Treml et~al.(2016)Treml, Arjona-Medina, Unterthiner, Durgesh,
  Friedmann, Schuberth, Mayr, Heusel, Hofmarcher, Widrich
  et~al.}]{treml2016speeding}
\bibinfo{author}{Treml, M.}, \bibinfo{author}{Arjona-Medina, J.},
  \bibinfo{author}{Unterthiner, T.}, \bibinfo{author}{Durgesh, R.},
  \bibinfo{author}{Friedmann, F.}, \bibinfo{author}{Schuberth, P.},
  \bibinfo{author}{Mayr, A.}, \bibinfo{author}{Heusel, M.},
  \bibinfo{author}{Hofmarcher, M.}, \bibinfo{author}{Widrich, M.} et~al.
  (\bibinfo{year}{2016}).
\newblock \bibinfo{title}{Speeding up semantic segmentation for autonomous
  driving}.
\newblock In {\it \bibinfo{booktitle}{MLITS, NIPS Workshop}\/}.
\newblock volume~\bibinfo{volume}{2}.
\bibitem[{Vaswani et~al.(2017)Vaswani, Shazeer, Parmar, Uszkoreit, Jones,
  Gomez, Kaiser \& Polosukhin}]{vaswani2017attention}
\bibinfo{author}{Vaswani, A.}, \bibinfo{author}{Shazeer, N.},
  \bibinfo{author}{Parmar, N.}, \bibinfo{author}{Uszkoreit, J.},
  \bibinfo{author}{Jones, L.}, \bibinfo{author}{Gomez, A.~N.},
  \bibinfo{author}{Kaiser, L.}, \& \bibinfo{author}{Polosukhin, I.}
  (\bibinfo{year}{2017}).
\newblock \bibinfo{title}{Attention is all you need}.
\newblock {\it \bibinfo{journal}{arXiv preprint arXiv:1706.03762}\/}, .
\bibitem[{Wang et~al.(2020{\natexlab{a}})Wang, Wang, Du, Yang, Zhang, Ding,
  Mardziel \& Hu}]{wang2020score}
\bibinfo{author}{Wang, H.}, \bibinfo{author}{Wang, Z.}, \bibinfo{author}{Du,
  M.}, \bibinfo{author}{Yang, F.}, \bibinfo{author}{Zhang, Z.},
  \bibinfo{author}{Ding, S.}, \bibinfo{author}{Mardziel, P.}, \&
  \bibinfo{author}{Hu, X.} (\bibinfo{year}{2020}{\natexlab{a}}).
\newblock \bibinfo{title}{Score-cam: Score-weighted visual explanations for
  convolutional neural networks}.
\newblock In {\it \bibinfo{booktitle}{Proceedings of the IEEE/CVF Conference on
  Computer Vision and Pattern Recognition Workshops}\/} (pp.
  \bibinfo{pages}{24--25}).
\bibitem[{Wang et~al.(2020{\natexlab{b}})Wang, Sun, Cheng, Jiang, Deng, Zhao,
  Liu, Mu, Tan, Wang et~al.}]{wang2020deep}
\bibinfo{author}{Wang, J.}, \bibinfo{author}{Sun, K.}, \bibinfo{author}{Cheng,
  T.}, \bibinfo{author}{Jiang, B.}, \bibinfo{author}{Deng, C.},
  \bibinfo{author}{Zhao, Y.}, \bibinfo{author}{Liu, D.}, \bibinfo{author}{Mu,
  Y.}, \bibinfo{author}{Tan, M.}, \bibinfo{author}{Wang, X.} et~al.
  (\bibinfo{year}{2020}{\natexlab{b}}).
\newblock \bibinfo{title}{Deep high-resolution representation learning for
  visual recognition}.
\newblock {\it \bibinfo{journal}{IEEE transactions on pattern analysis and
  machine intelligence}\/}, .
\bibitem[{Wang et~al.(2020{\natexlab{c}})Wang, Wu, Zhu, Li, Zuo \&
  Hu}]{wang2020eca}
\bibinfo{author}{Wang, Q.}, \bibinfo{author}{Wu, B.}, \bibinfo{author}{Zhu,
  P.}, \bibinfo{author}{Li, P.}, \bibinfo{author}{Zuo, W.}, \&
  \bibinfo{author}{Hu, Q.} (\bibinfo{year}{2020}{\natexlab{c}}).
\newblock \bibinfo{title}{Eca-net: efficient channel attention for deep
  convolutional neural networks, 2020 ieee}.
\newblock In {\it \bibinfo{booktitle}{CVF Conference on Computer Vision and
  Pattern Recognition (CVPR). IEEE}\/}.
\bibitem[{Woo et~al.(2018)Woo, Park, Lee \& Kweon}]{woo2018cbam}
\bibinfo{author}{Woo, S.}, \bibinfo{author}{Park, J.}, \bibinfo{author}{Lee,
  J.-Y.}, \& \bibinfo{author}{Kweon, I.~S.} (\bibinfo{year}{2018}).
\newblock \bibinfo{title}{Cbam: Convolutional block attention module}.
\newblock In {\it \bibinfo{booktitle}{Proceedings of the European conference on
  computer vision (ECCV)}\/} (pp. \bibinfo{pages}{3--19}).
\bibitem[{Xie et~al.(2019)Xie, Wu, Maaten, Yuille \& He}]{xie2019feature}
\bibinfo{author}{Xie, C.}, \bibinfo{author}{Wu, Y.}, \bibinfo{author}{Maaten,
  L. v.~d.}, \bibinfo{author}{Yuille, A.~L.}, \& \bibinfo{author}{He, K.}
  (\bibinfo{year}{2019}).
\newblock \bibinfo{title}{Feature denoising for improving adversarial
  robustness}.
\newblock In {\it \bibinfo{booktitle}{Proceedings of the IEEE/CVF Conference on
  Computer Vision and Pattern Recognition}\/} (pp. \bibinfo{pages}{501--509}).
\bibitem[{Yu et~al.(2020{\natexlab{a}})Yu, Gao, Wang, Yu, Shen \&
  Sang}]{yu2020bisenet}
\bibinfo{author}{Yu, C.}, \bibinfo{author}{Gao, C.}, \bibinfo{author}{Wang,
  J.}, \bibinfo{author}{Yu, G.}, \bibinfo{author}{Shen, C.}, \&
  \bibinfo{author}{Sang, N.} (\bibinfo{year}{2020}{\natexlab{a}}).
\newblock \bibinfo{title}{Bisenet v2: Bilateral network with guided aggregation
  for real-time semantic segmentation}.
\newblock {\it \bibinfo{journal}{arXiv preprint arXiv:2004.02147}\/}, .
\bibitem[{Yu et~al.(2020{\natexlab{b}})Yu, Wang, Gao, Yu, Shen \&
  Sang}]{yu2020context}
\bibinfo{author}{Yu, C.}, \bibinfo{author}{Wang, J.}, \bibinfo{author}{Gao,
  C.}, \bibinfo{author}{Yu, G.}, \bibinfo{author}{Shen, C.}, \&
  \bibinfo{author}{Sang, N.} (\bibinfo{year}{2020}{\natexlab{b}}).
\newblock \bibinfo{title}{Context prior for scene segmentation}.
\newblock In {\it \bibinfo{booktitle}{Proceedings of the IEEE/CVF Conference on
  Computer Vision and Pattern Recognition}\/} (pp.
  \bibinfo{pages}{12416--12425}).
\bibitem[{Yu et~al.(2018{\natexlab{a}})Yu, Wang, Peng, Gao, Yu \&
  Sang}]{yu2018bisenet}
\bibinfo{author}{Yu, C.}, \bibinfo{author}{Wang, J.}, \bibinfo{author}{Peng,
  C.}, \bibinfo{author}{Gao, C.}, \bibinfo{author}{Yu, G.}, \&
  \bibinfo{author}{Sang, N.} (\bibinfo{year}{2018}{\natexlab{a}}).
\newblock \bibinfo{title}{Bisenet: Bilateral segmentation network for real-time
  semantic segmentation}.
\newblock In {\it \bibinfo{booktitle}{Proceedings of the European conference on
  computer vision (ECCV)}\/} (pp. \bibinfo{pages}{325--341}).
\bibitem[{Yu et~al.(2018{\natexlab{b}})Yu, Wang, Peng, Gao, Yu \&
  Sang}]{yu2018learning}
\bibinfo{author}{Yu, C.}, \bibinfo{author}{Wang, J.}, \bibinfo{author}{Peng,
  C.}, \bibinfo{author}{Gao, C.}, \bibinfo{author}{Yu, G.}, \&
  \bibinfo{author}{Sang, N.} (\bibinfo{year}{2018}{\natexlab{b}}).
\newblock \bibinfo{title}{Learning a discriminative feature network for
  semantic segmentation}.
\newblock In {\it \bibinfo{booktitle}{Proceedings of the IEEE conference on
  computer vision and pattern recognition}\/} (pp.
  \bibinfo{pages}{1857--1866}).
\bibitem[{Yu et~al.(2021)Yu, Xiao, Gao, Yuan, Zhang, Sang \& Wang}]{yu2021lite}
\bibinfo{author}{Yu, C.}, \bibinfo{author}{Xiao, B.}, \bibinfo{author}{Gao,
  C.}, \bibinfo{author}{Yuan, L.}, \bibinfo{author}{Zhang, L.},
  \bibinfo{author}{Sang, N.}, \& \bibinfo{author}{Wang, J.}
  (\bibinfo{year}{2021}).
\newblock \bibinfo{title}{Lite-hrnet: A lightweight high-resolution network}.
\newblock {\it \bibinfo{journal}{arXiv preprint arXiv:2104.06403}\/}, .
\bibitem[{Yuan et~al.(2019)Yuan, Chen \& Wang}]{yuan2019object}
\bibinfo{author}{Yuan, Y.}, \bibinfo{author}{Chen, X.}, \&
  \bibinfo{author}{Wang, J.} (\bibinfo{year}{2019}).
\newblock \bibinfo{title}{Object-contextual representations for semantic
  segmentation}.
\newblock {\it \bibinfo{journal}{arXiv preprint arXiv:1909.11065}\/}, .
\bibitem[{Zhang et~al.(2020)Zhang, Wu, Zhang, Zhu, Lin, Zhang, Sun, He,
  Mueller, Manmatha et~al.}]{zhang2020resnest}
\bibinfo{author}{Zhang, H.}, \bibinfo{author}{Wu, C.}, \bibinfo{author}{Zhang,
  Z.}, \bibinfo{author}{Zhu, Y.}, \bibinfo{author}{Lin, H.},
  \bibinfo{author}{Zhang, Z.}, \bibinfo{author}{Sun, Y.}, \bibinfo{author}{He,
  T.}, \bibinfo{author}{Mueller, J.}, \bibinfo{author}{Manmatha, R.} et~al.
  (\bibinfo{year}{2020}).
\newblock \bibinfo{title}{Resnest: Split-attention networks}.
\newblock {\it \bibinfo{journal}{arXiv preprint arXiv:2004.08955}\/}, .
\bibitem[{Zhao et~al.(2018)Zhao, Qi, Shen, Shi \& Jia}]{zhao2018icnet}
\bibinfo{author}{Zhao, H.}, \bibinfo{author}{Qi, X.}, \bibinfo{author}{Shen,
  X.}, \bibinfo{author}{Shi, J.}, \& \bibinfo{author}{Jia, J.}
  (\bibinfo{year}{2018}).
\newblock \bibinfo{title}{Icnet for real-time semantic segmentation on
  high-resolution images}.
\newblock In {\it \bibinfo{booktitle}{Proceedings of the European conference on
  computer vision (ECCV)}\/} (pp. \bibinfo{pages}{405--420}).
\bibitem[{Zhao et~al.(2017)Zhao, Shi, Qi, Wang \& Jia}]{zhao2017pyramid}
\bibinfo{author}{Zhao, H.}, \bibinfo{author}{Shi, J.}, \bibinfo{author}{Qi,
  X.}, \bibinfo{author}{Wang, X.}, \& \bibinfo{author}{Jia, J.}
  (\bibinfo{year}{2017}).
\newblock \bibinfo{title}{Pyramid scene parsing network}.
\newblock In {\it \bibinfo{booktitle}{Proceedings of the IEEE conference on
  computer vision and pattern recognition}\/} (pp.
  \bibinfo{pages}{2881--2890}).
\bibitem[{Zhuang et~al.(2019)Zhuang, Yang, Gu \& Dvornek}]{zhuang2019shelfnet}
\bibinfo{author}{Zhuang, J.}, \bibinfo{author}{Yang, J.}, \bibinfo{author}{Gu,
  L.}, \& \bibinfo{author}{Dvornek, N.} (\bibinfo{year}{2019}).
\newblock \bibinfo{title}{Shelfnet for fast semantic segmentation}.
\newblock In {\it \bibinfo{booktitle}{Proceedings of the IEEE/CVF International
  Conference on Computer Vision Workshops}\/} (pp. \bibinfo{pages}{0--0}).

\end{thebibliography}

\end{document}